\documentclass[journal,transmag]{IEEEtran}

\ifCLASSINFOpdf

\else

\fi

\begin{document}
\bibliographystyle{unsrt}
\title{A Survey of Simultaneous Localization and Mapping with an Envision in 6G Wireless Networks}


\author{\IEEEauthorblockN{Baichuan Huang\IEEEauthorrefmark{1,2},
Jun Zhao\IEEEauthorrefmark{1}
Jingbin Liu\IEEEauthorrefmark{2}}
\IEEEauthorblockA{\IEEEauthorrefmark{1}Nanyang Technological University, Singapore}
\IEEEauthorblockA{\IEEEauthorrefmark{2}Wuhan University, Wuhan, China}
\IEEEauthorblockA{\IEEEauthorrefmark{}(huangbaichuan@whu.edu.cn)}
}


\IEEEtitleabstractindextext{%
\begin{abstract}
Simultaneous Localization and Mapping (SLAM) achieves the purpose of simultaneous positioning and map construction based on self-perception. The paper makes an overview in SLAM including Lidar SLAM, visual SLAM, and their fusion. For Lidar or visual SLAM, the survey illustrates the basic type and product of sensors, open source system in sort and history, deep learning embedded, the challenge and future. Additionally, visual inertial odometry is supplemented. For Lidar and visual fused SLAM, the paper highlights the multi-sensors calibration, the fusion in hardware, data, task layer. The open question and forward thinking with an envision in 6G wireless networks end the paper. The contributions of this paper can be summarized as follows: the paper provides a high quality and full-scale overview in SLAM. It’s very friendly for new researchers to hold the development of SLAM and learn it very obviously. Also, the paper can be considered as dictionary for experienced researchers to search and find new interested orientation.  
\end{abstract}

}

\maketitle

\IEEEdisplaynontitleabstractindextext

\IEEEpeerreviewmaketitle

\section{Introduction}

\IEEEPARstart{S}{lam} is the abbreviation of Simultaneous Localization and Mapping, which contains two main tasks, localization and mapping. It is a significant open problem in mobile robotics: to move precisely, a mobile robot must have an accurate environment map; however, to build an accurate map, the mobile robot’s sensing locations must be known precisely \cite{leonard1991simultaneous}. In this way, simultaneous map building and localization can be seen to present a question of “which came first, the chicken or the egg?” (The map or the motion?)

In 1990, \cite{smith1990estimating} firstly proposed the use of the EKF (Extended Kalman Filter) for incrementally estimating the posterior distribution over robot pose along with the positions of the landmarks. In fact, starting from the unknown location of the unknown environment, the robot locates its own position and attitude through repeated observation of environmental features in the movement process, and then builds an incremental map of the surrounding environment according to its own position, so as to achieve the purpose of simultaneous positioning and map construction. Localization is a very complex and hot point in recent years. The technologies of localization depend on environment and demand for cost, accuracy, frequency and robustness, which can be achieved by GPS (Global Positioning System), IMU (Inertial Measurement Unit), and wireless signal, etc.\cite{huang2019robust}\cite{liu2012iparking}. But GPS can only work well outdoors and IMU system has cumulative error \cite{liu2012hybrid}. The technology of wireless, as an active system, can't make a balance between cost and accuracy. With the fast development, SLAM equipped with Lidar, camera, IMU and other sensors springs up in last years.

Begin with filter-based SLAM, Graph-based SLAM play a dominant role now. The algorithm derives from KF (Kalman Filter), EKF and PF (Particle Filter) to graph-based optimization. And single thread has been replaced by multi-thread. The technology of SLAM also changed from the earliest prototype of military use to later robot applications with the fusion of multi sensors.

The organization of this paper can be summarized as follows: in Section II, Lidar SLAM including Lidar sensors, open source Lidar SLAM system, deep learning in Lidar and challenge as well as future will be illustrated. Section III highlights the visual SLAM including camera sensors, different density of open source visual SLAM system, visual inertial odometry SLAM, deep learning in visual SLAM and future. In Section IV, the fusion of Lidar and vision will be demonstrated, with several directions for future research of SLAM. Section V illustrates the envision in combination of 6G and SLAM. Conclusion is put in section VI. The paper provides high quality and full-scale user guide for new researchers in SLAM.
\section{Lidar SLAM}
In 1991, \cite{leonard1991simultaneous} used multiple servo-mounted sonar sensors and EKF filter to equip robots with SLAM system. Begin with sonar sensors, the birth of Lidar makes SLAM system more reliable and robustness.
\subsection{Lidar Sensors}
Lidar sensors can be divided into 2D Lidar and 3D Lidar, which are defined by the number of Lidar beams. In terms of production process, Lidar can also be divided into mechanical Lidar, hybrid solid-state Lidar like MEMS (micro-electro-mechanical) and solid-state Lidar. Solid-state Lidar can be produced by the technology of phased array and flash.
\begin{itemize}
    \item \textbf{Velodyne}: In mechanical Lidar, it has VLP-16, HDL-32E and HDL-64E. In hybrid solid-state Lidar, it has Ultra puck auto with 32E.  
    \item \textbf{SLAMTEC}: it has low cost Lidar and robot platform such RPLIDAR A1, A2 and R3. 
    \item \textbf{Ouster}: it has mechanical Lidar from 16 to 128 channels.
    \item \textbf{Quanergy}: S3 is the first issued solid-state Lidar in the world and M8 is the mechanical Lidar. The S3-QI is the micro solid-state Lidar.
    \item \textbf{Ibeo}: It has Lux 4L and Lux 8L in mechanical Lidar. Cooperated with Valeo, it issued a hybrid solid-state Lidar named Scala.
\end{itemize}

In the trend, miniaturization and lightweight solid state Lidar will occupied the market and be satisfied with most application. Other Lidar companies include but not limited to \textbf{sick}, \textbf{Hokuyo}, \textbf{HESAI}, \textbf{RoboSense}, \textbf{LeddarTech}, \textbf{ISureStar}, \textbf{benewake}, \textbf{Livox}, \textbf{Innovusion}, \textbf{Innoviz}, \textbf{Trimble}, \textbf{Leishen Intelligent System}.
\subsection{Lidar SLAM System}
Lidar SLAM system is reliable in theory and technology. \cite{thrun2005probabilistic} illustrated the theory in math about how to simultaneous localization and mapping with 2D Lidar based on probabilistic. Furthre, \cite{santos2013evaluation} make surveys about 2D Lidar SLAM system. 
\subsubsection{2D SLAM}
\begin{itemize}
    \item \textbf{Gmapping}: it is the most used SLAM package in robots based on RBPF (Rao-Blackwellisation Partical Filter) method. It adds scan-match method to estimate the position\cite{grisetti2007improved}\cite{thrun2005probabilistic}. It is the improved version with Grid map based on \textbf{FastSLAM}\cite{montemerlo2002fastslam}\cite{montemerlo2003fastslam}.
    \item \textbf{HectorSlam}: it combines a 2D SLAM system  and 3D navigation with scan-match technology and an inertial sensing system\cite{kohlbrecher2011flexible}.
    \item \textbf{KartoSLAM}: it is a graph-based SLAM system\cite{konolige2010efficient}. 
    \item \textbf{LagoSLAM}: its basic is the graph-based SLAM, which is the minimization of a nonlinear non-convex cost function\cite{carlone2012linear}. 
    \item \textbf{CoreSLAm}: it is an algorithm to be understood with minimum loss of performance\cite{steux2010slam}. 
    \item \textbf{Cartographer}: it is a SLAM system from Google\cite{hess2016real}. It adopted sub-map and loop closure to achieve a better performance in product grade. The algorithm can provide SLAM in 2D and 3D across multiple platforms and sensor configurations.
\end{itemize}
\subsubsection{3D SLAM}
\begin{itemize}
    \item \textbf{Loam}: it is a real-time method for state estimation and mapping using a 3D Lidar\cite{zhang2014loam}. It also has back and forth spin version and continuous scanning 2D Lidar version.
    \item \textbf{Lego-Loam}: it takes in point cloud from a Velodyne VLP-16 Lidar (placed horizontal) and optional IMU data as inputs. The system outputs 6D pose estimation in real-time and has global optimization and loop closure\cite{shan2018lego}.
    \item \textbf{Cartographer}: it supports 2D and 3D SLAM\cite{hess2016real}.
    \item \textbf{IMLS-SLAM}: it presents a new low-drift SLAM algorithm based only on 3D LiDAR data based on a scan-to-model matching framework \cite{deschaud2018imls}.
\end{itemize}

\subsubsection{Deep Learning With Lidar SLAM}
\begin{itemize}
    \item \textbf{Feature \& Detection}: \textbf{PointNetVLAD} \cite{angelina2018pointnetvlad} allows end-to-end training and inference to extract the global descriptor from a given 3D point cloud to solve point cloud based retrieval for place recognition. \textbf{VoxelNet} \cite{zhou2018voxelnet} is a generic 3D detection network that unifies feature extraction and bounding box prediction into a single stage, end-to-end trainable deep network. Other work can be seen in \textbf{BirdNet} \cite{beltran2018birdnet}. \textbf{LMNet} \cite{minemura2018lmnet}  describes an efficient single-stage deep convolutional neural network to detect objects and outputs an objectness map and the bounding box offset values for each point. \textbf{PIXOR} \cite{yang2018pixor} is a proposal-free, single-stage detector that outputs oriented 3D object estimates decoded from pixel-wise neural network predictions. \textbf{Yolo3D} \cite{ali2018yolo3d} builds on the success of the one-shot regression meta-architecture in the 2D perspective image space and extend it to generate oriented 3D object bounding boxes from LiDAR point cloud. . \textbf{PointCNN} \cite{li2018pointcnn} proposes to learn a X-transformation from the input points. The X-transformation is applied by element-wise product and sum operations of typical convolution operator. \textbf{MV3D} \cite{chen2017multi} is a sensory-fusion framework that takes both Lidar point cloud and RGB images as input and predicts oriented 3D bounding boxes. \textbf{PU-GAN} \cite{li2019pu} presents a new point cloud upsampling network based on a generative adversarial network (GAN). Other similar work can be seen in this best paper in CVPR2018 but not limited to \cite{su2018splatnet}. 
    \item \textbf{Recognition \& Segmentation}: In fact, the method of segmentation to 3D point cloud can be divided into Edge-based, region growing, model fitting, hybrid method, machine learning application and deep learning \cite{grilli2017review}. Here the paper focuses on the methods of deep learning. \textbf{PointNet} \cite{qi2016pointnet} designs a novel type of neural network that directly consumes point clouds, which has the function of classification, segmentation and semantic analysis. \textbf{PointNet++} \cite{qi2017pointnetplusplus} learns hierarchical features with increasing scales of contexts. \textbf{VoteNet} \cite{qi2019deep} constructs a 3D detection pipeline for point cloud as a end-to-end 3D object detection network, which is based on PointNet++. \textbf{SegMap} \cite{segmap2018} is a map representation solution to the localization and mapping problem based on the extraction of segments in 3D point clouds. \textbf{SqueezeSeg} \cite{wu2017squeezeseg}\cite{wu2018squeezesegv2}\cite{yue2018lidar} are convolutional neural nets with recurrent CRF (Conditional random fields) for real-time road-object segmentation from 3d Lidar point cloud. \textbf{PointSIFT} \cite{jiang2018pointsift} is a semantic segmentation framework for 3D point clouds. It is based on a simple module which extracts features from neighbor points in eight directions. \textbf{PointWise} \cite{hua-pointwise-cvpr18} presents a convolutional neural network for semantic segmentation and object recognition with 3D point clouds. \textbf{3P-RNN} \cite{ye20183d} is a novel end-to-end approach for unstructured point cloud semantic segmentation along two horizontal directions to exploit the inherent contextual features. Other similar work can be seen but not limited to \textbf{SPG} \cite{landrieu2018large} and the review \cite{grilli2017review}. \textbf{SegMatch} \cite{dube2017segmatch} is a loop closure method based on the detection and matching of 3D segments. \textbf{Kd-Network}  \cite{klokov2017escape} is designed for 3D model recognition tasks and works with unstructured point clouds. \textbf{DeepTemporalSeg} \cite{dewan2019deeptemporalseg} propose a deep convolutional neural network (DCNN) for the semantic segmentation of a LiDAR scan with temporally consistency. \textbf{LU-Net} \cite{LU-Net2019} achieve the function of semantic segmentation instead of applying some global 3D segmentation method. Other similar work can be seen but not limited to \textbf{PointRCNN} \cite{shi2019pointrcnn}.
    \item \textbf{Localization}:  \textbf{L3-Net} \cite{L3-Net} is a novel learning-based LiDAR localization system that achieves centimeter-level localization accuracy. \textbf{SuMa++} \cite{Chen2019suma} computes semantic segmentation results in point-wise labels for the whole scan, allowing us to build a semantically-enriched map with labeled surfels and  improve the projective scan matching via semantic constraints.
\end{itemize}

\subsection{Challenge and Future}
\subsubsection{Cost and Adaptability}
The advantage of Lidar is that it can provide 3D information, and it is not affected by night and light change. In addition, the angle of view is relatively large and can reach 360 degrees.
But the technological threshold of Lidar is very high, which lead to long development cycle and unaffordable cost on a large scale. In the future, miniaturization, reasonable cost, solid state, and achieving high reliability and adaptability is the trend.
\subsubsection{Low-Texture and Dynamic Environment}
Most SLAM system can just work in a fixed environment but things change constantly. Besides, low-Texture environment like long corridor and big pipeline will make trouble for Lidar SLAM. 
\cite{wang2018imu} uses IMU to assist 2D SLAM to solve above obstacles. Further, \cite{walcott2012dynamic} incorporates the time dimension into the mapping process to enable a robot to maintain an accurate map while operating in dynamical environments. How to make Lidar SLAM more robust to low-texture and dynamic environment, and how to keep map updated should be taken into consideration more deeply.
\subsubsection{Adversarial Sensor Attack}
Deep Neural Network is easily attacked by adversarial samples, which is also proved in  camera-based perception. But in Lidar-based perception, it is highly important but unexplored. By relaying attack, \cite{shin2017illusion} firstly spoofs the Lidar with interference in output data and distance estimation. The novel saturation attack completely incapacitate a Lidar from sensing a certain direction based on Velodyne’s VLP-16. \cite{cao2019adversarial} explores the possibility of strategically controlling the spoofed attack to fool the machine learning model. The paper regards task as an optimization problem and design modeling methods for the input perturbation function and the objective function., which improves the attack success rates to around 75\%. The adversarial sensor attack will spoof the SLAM system based on Lidar point cloud, which is invisible as hardly found and defended. In the case, research on how to prevent the Lidar SLAM system from adversarial sensor attack should be a new topic.

\section{Visual SLAM}
As the development of CPU and GPU, the capability of graphics processing  becomes more and more powerful. Camera sensors getting cheaper, more lightweight and more versatile at the same time. The past decade has seen the rapid development of visual SLAM. Visual SLAM using camera also make the system cheaper and smaller compare with Lidar system. Now, visual SLAM system can run in micro PC and embedded device, even in mobile devices like smart phones \cite{mur2015orb}\cite{qin2018vins}\cite{klein2009parallel}\cite{InfiniTAM_ISMAR_2015}\cite{lynen2015get}. 

Visual SLAM includes collection of sensors' data such as camera or inertial measurement unit , Visual Odometry or Visual Inertial Odometry in front end, Optimization in back end, Loop closure in back end and Mapping \cite{Gao2017SLAM}.  Relocalization is the additional modules for stable and accurate visual SLAM \cite{taketomi2017visual}. 

In process of Visual Odometry, in addition to the method based on features or template matching, or correlation methods to determine the motion of the camera, there is another method relying on the Fourier-Mellin Transform \cite{reddy1996fft}. \cite{kazik2011visual} and \cite{birem2018visual} give the example in the environment with  no distinct visual features when use the ground-facing camera.

\subsection{Visual Sensors}
The most used sensors that visual SLAM based are cameras. In detail, camera can be divided into monocular camera, stereo camera, RGB-D camera, event camera, etc.

\textbf{Monocular camera}: visual slam based on monocular camera have a scale with real size of track and map. That's say that the real depth can't be got by monocular camera, which called Scale Ambiguity \cite{Zhangguofeng2016}. The SLAM based on Monocular camera has to initialization, and face the problem of drift.

\textbf{Stereo camera}: stereo camera is a combination of two monocular camera but the distance called baseline between the two monocular camera is known. Although the depth can be got based on calibration, correction, matching and calculation, the process will be a waste of lost of resources. 

\textbf{RGB-D camera}: RGB-D camera also called depth camera because the camera can output depth in pixel directly. The depth camera can be realized by technology of stereo, structure-light and TOF. The theory of Structure-light is that infrared laser emits some pattern with structure feature to the surface of object. Then the IR camera will collect the change of patter due to the different depth in the surface. TOF will measure the time of laser's flight to calculate the distance.

\textbf{Event camera}: \cite{Gallego2019Event} illustrates that instead of capturing images at a fixed rate, event camera measures per-pixel brightness changes asynchronously. Event camera has very high dynamic range (140 dB vs. 60 dB), high temporal resolution (in the order of us), low power consumption, and do not suffer from motion blur. Hence, event cameras can performance better than traditional camera in high speed and high dynamic range. The example of the event camera are Dynamic Vision Sensor \cite{lichtsteiner2008128}\cite{son20174}\cite{posch2009microbolometer}\cite{hofstatter2010sparc}, Dynamic Line Sensor \cite{posch2007dual}, Dynamic and Active-Pixel Vision Sensor \cite{brandli2014240}, and Asynchronous Time-based Image Sensor \cite{posch2010qvga}.

Next the product and company of visual sensors will be introduced:

\begin{itemize}
    \item \textbf{Microsoft}: Kinectc v1(structured-light), Kinect v2(TOF), Azure Kinect(with microphone and IMU).
    \item \textbf{Intel}: 200 Series, 300 Series, Module D400 Series, D415(Active IR Stereo, Rolling shutter), D435(Active IR Stereo, Global Shutter), D435i(D435 with IMU).
    \item \textbf{Stereolabs ZED}: ZED Stereo camera(depth up to 20m).
    \item \textbf{MYNTAI}: D1000 Series(depth camera), D1200(for smart phone), S1030 Series(standard stereo camera).
    \item \textbf{Occipital Structure}: Structure Sensor(Suitable for ipad).
    \item \textbf{Samsung}: Gen2 and Gen3 dynamic vision sensors and event-based vision solution\cite{son20174}.
\end{itemize}

Other depth camera can be listed as follows but not limited to \textbf{Leap Motion}, \textbf{Orbbec Astra}, \textbf{Pico Zense}, \textbf{DUO}, \textbf{Xtion}, \textbf{Camboard}, \textbf{IMI}, \textbf{Humanplus}, \textbf{PERCIPIO.XYZ}, \textbf{PrimeSense}. Other event camera can be listed as follows but not limited to \textbf{iniVation}, \textbf{AIT(AIT Austrian Institute of Technology)}, \textbf{SiliconEye}, \textbf{Prophesee}, \textbf{CelePixel}, \textbf{Dilusense}.

\subsection{Visual SLAM System}
The method of utilizing information from image can be classified into direct method and feature based method. Direct method leads to semiDense and dense construction while feature based method cause sparse construction. Next, some visual slam will be introduced ( ATAM7 is a visual SLAM toolkit for beginners\cite{taketomi2017visual}): 

\subsubsection{Sparse Visual SLAM}
\begin{itemize}
    \item \textbf{MonoSLAM}: it (monocular) is the first real-time mono SLAM system, which is based on EKF\cite{davison2007monoslam}. 
    \item \textbf{PTAM}: it (monocular) is the first SLAM system that parallel tracking and mapping. It firstly adopts Bundle Adjustment to optimize and concept of key frame \cite{klein2007parallel}\cite{klein2009parallel}. The later version supports a trivially simple yet effective relocalization method \cite{klein2008improving}.
    \item \textbf{ORB-SLAM}: it (monocular) uses three threads: Tracking, Local Mapping and Loop Closing \cite{rublee2011orb}\cite{mur2015orb}. \textbf{ORB-SLAM v2} \cite{mur2017orb} supports monocular, stereo, and RGB-D cameras. \textbf{CubemapSLAM} \cite{wang2018cubemapslam} is a  SLAM system for monocular fisheye cameras based on ORB-SLAM. \textbf{Visual Inertial ORB-SLAM} \cite{mur2017visual}\cite{forster2016manifold} explains the initialization process of IMU and the joint optimization with visual information.
    \item \textbf{proSLAM}: it (stereo) is a lightweight visual SLAM system with easily understanding \cite{2018-schlegel-proslam}.
    \item \textbf{ENFT-sfm}: it (monocular) is a feature tracking method which can efficiently match feature point correspondences among one or multiple video sequences \cite{zhang2016efficient}. The updated version \textbf{ENFT-SLAM} can run in large scale.
    \item \textbf{OpenVSLAm}: it (all types of cameras) \cite{openvslam2019} is based on an indirect SLAM algorithm with sparse features. The excellent point of OpenVSLAM is that the system supports perspective, fisheye, and equirectangular, even the camera models you design.
    \item \textbf{TagSLAM}: it realizes SLAM with AprilTag fiducial markers \cite{pfrommer2019tagslam}. Also, it provides a front end to the GTSAM factor graph optimizer, which can design lots of experiments.
\end{itemize}

Other similar work can be listed as follows but not limited to \textbf{UcoSLAM} \cite{munoz2019ucoslam}.

\subsubsection{SemiDense Visual SLAM}

\begin{itemize}
    \item \textbf{LSD-SLAM}:  it (monocular) proposes a novel direct tracking method which operates on Lie Algebra and direct method \cite{engel2014lsd}. \cite{engel2015large} make it supporting stereo cameras and \cite{caruso2015large} make it supporting omnidirectional cameras. Other similar work with omnidirectional cameras can be seen in \cite{li2018spherical}.
    \item \textbf{SVO}: it (monocular) is Semi-direct Visual Odoemtry \cite{forster2016svo}. It uses sparse model-based image alignment to get a fast speed. The update version is extended to multiple cameras, fisheye and catadioptric ones \cite{forster2016manifold}. \cite{forster2016manifold} gives detailed math proof about VIO. \textbf{CNN-SVO} \cite{loo2018cnn} is the version of  SVO with the depth prediction from a single-image depth prediction network.
    \item \textbf{DSO}:  it (monocular) \cite{DBLP:journals/corr/EngelKC16}\cite{engel2017direct} is a new work from the author of LSD-SLAM \cite{engel2014lsd}. The work creates a visual odoemtry based on direct method and sparse method without detection and description of feature point.
    \item \textbf{EVO}: it (Event camera) \cite{rebecq2016evo} is an event-based visual odometry algorithm. Our algorithm is unaffected by motion blur and operates very well in challenging, high dynamic range conditions with strong illumination changes. Other semiDense SLAM based on event camera can be seen in \cite{zhou2018semi}. Other VO (visual odometry) system based on event camera can be seen in \cite{weikersdorfer2013simultaneous}\cite{weikersdorfer2014event}.

\end{itemize}

\subsubsection{Dense Visual SLAM}

\begin{itemize}
    \item \textbf{DTAM}:  it (monocular) can reconstruct 3D model in real time based on minimizing a global spatially regularized energy functional in a novel non-convex optimization framework, which is called direct method \cite{civera2008inverse}\cite{newcombe2011dtam}.
    \item \textbf{MLM SLAM}: it (monocular) can reconstruct dense 3D model online without graphics processing unit (GPU) \cite{greene2016multi}. The key contribution is a multi-resolution depth estimation and spatial smoothing process.
    \item \textbf{Kinect Fusion}: it (RGB-D) is almost the first 3D reconstruction system with depth camera \cite{newcombe2011kinectfusion}\cite{izadi2011kinectfusion}.
    \item \textbf{DVO}: it (RGB-D)  proposes a dense visual SLAM method, an entropy-based similarity measure for keyframe selection and loop closure detection based g2o framework \cite{steinbrucker2011real}\cite{kerl2013robust}\cite{kerl2013dense}.
    \item \textbf{RGBD-SLAM-V2}: it (RGB-D) can reconstruct accurate 3D dense model without the help of other sensors. \cite{endres20133}.
    \item \textbf{Kintinuous}: it (RGB-D) is a visual SLAM system with globally consistent point and mesh reconstructions in real-time \cite{whelan2012kintinuous}\cite{whelan2015real}\cite{Whelan2011Robust}.
    \item \textbf{RTAB-MAP}: it (RGB-D) supports simultaneous localization and mapping but it's hard to be basis to develop upper algorithm \cite{labbe2014online}\cite{labbeappearance}\cite{labbe2011memory}. The latter version support both visual and Lidar SLAM \cite{labbe2019rtab}.
    \item \textbf{Dynamic Fusion}: it (RGB-D) presents the first dense SLAM system capable of reconstructing non-rigidly deforming scenes in real-time based Kinect Fusion \cite{newcombe2015dynamicfusion}. \textbf{VolumeDeform} \cite{innmann2016volumedeform} also realizes real-time non-rigid reconstruction but not open source. The similar work can be seen in \textbf{Fusion4D} \cite{dou2016fusion4d}.
    \item \textbf{Elastic Fusion}: it (RGB-D) is a real-time dense visual SLAM system capable of capturing comprehensive dense globally consistent surfel-based maps of room scale environments explored using an RGB-D camera \cite{whelan2015elasticfusion}\cite{whelan2016elasticfusion}.
    \item \textbf{InfiniTAM}: it (RGB-D) is a real time 3D reconstruction system with CPU in Linux, IOS, Android platform \cite{InfiniTAM_ISMAR_2015}\cite{InfiniTAM_arXiv_2017}\cite{InfiniTAM_ECCV_2016}.
    \item \textbf{Bundle Fusion}: it (RGB-D) supports robust tracking with recovery from gross tracking failures and re-estimates the 3D model in real-time to ensure global consistency \cite{dai2017bundlefusion}.
    \item \textbf{KO-Fusion}: it (RGB-D) \cite{houseago2019ko} proposes a dense RGB-D SLAM system with kinematic and odometry measurements from a wheeled robot.
    \item \textbf{SOFT-SLAM}: it (stereo) \cite{cvivsic2017soft}  can create dense map with the advantages of large loop closing, which is based on SOFT \cite{cvivsic2015stereo} for pose estimation.
\end{itemize}

Other works can be listed as follows but not limited to \textbf{SLAMRecon}, \textbf{RKD-SLAM} \cite{liu2017robust} and \textbf{RGB-D SLAM} \cite{dai2018rgb}. \textbf{Maplab} \cite{schneider2018maplab}, \textbf{PointNVSNet} \cite{2019PointMVSNet},  \textbf{MID-Fusion}\cite{xu2018mid} and \textbf{MaskFusion} \cite{8613746} will introduced in next chapter. 
\subsubsection{Visual Inertial Odometry SLAM}
The determination of visual slam is technically challenging. Monocular visual SLAM has problems such as necessary initialization, scale ambiguity and scale drift \cite{strasdat2010scale}. Although stereo camera and RGB-D camera can solve the problems of initialization and scale, some obstacles can't be ignored such as fast movement (solved with Global Shuttle or fisheye even panoramic camera), small field of view, large calculation, occlusion, feature loss, dynamic scenes and changing light. Recently, VIO (visual inertial odometry SLAM) becomes the popular research. 

First of all, \cite{leutenegger2015keyframe}\cite{huang2014towards}\cite{li2013high} start some try in VIO. \cite{mur2017visual}\cite{forster2016manifold} give the samples and math proof in visual-inertial odeometry. \cite{Campos2019FastAR} use several rounds of visual-inertial bundle adjustment to make a robust initialization for VIO. Specially, tango \cite{froehlich2017investigation}, Dyson 360 Eye and hololens \cite{garon2016real} are the real products of VIO and receive good feedback. In addition to this
, ARkit (filter-based) from Apple, ARcore (filter-based) from Google, Inside-out from uSens are the technology of VIO. \textbf{PennCOSYVIO} \cite{pfrommer2017penncosyvio} synchronizes data from a VI-sensor (stereo camera and IMU), two Project Tango hand-held devices, and three GoPro Hero 4 cameras and calibrates intrinsically and extrinsically. Next some open source VIO system will be introduced \cite{delmerico2018benchmark}:
\begin{itemize}
    \item \textbf{SSF}: it (loosely-coupled, filter-based) is a time delay compensated single and multi sensor fusion framework based on an EKF \cite{weiss2012vision}.
    \item \textbf{MSCKF}: it (tightly-coupled, filter-based) is adopted by Google Tango based on extended Kalman filter \cite{mourikis2007multi}. But the similar work called \textbf{MSCKF-VIO} \cite{sun2018robust} open the source.
    \item \textbf{ROVIO}: it (tightly-coupled, filter-based) is an extended Kalman Filter with tracking of both 3D landmarks and image patch features \cite{bloesch2015robust}. It supports monocular camera.
    \item \textbf{OKVIS}: it (tightly-coupled, optimization-based) is an open and classic Keyframe-based Visual-Inertial SLAM \cite{leutenegger2015keyframe}. It supports monocular and stereo camera based sliding window estimator.
    \item \textbf{VINS}: \textbf{VINS-Mono} (tightly-coupled, optimization-based) \cite{li2017monocular}\cite{qin2018vins}\cite{qin2018online} is a real-time SLAM framework for Monocular Visual-Inertial Systems. The open source code runs on Linux, and is fully integrated with ROS.  \textbf{VINS-Mobile} \cite{qin2017robust}\cite{yang2016monocular} is a real-time monocular visual-inertial odometry running on compatible iOS devices. Furthermore, \textbf{VINS-Fusion} supports multiple visual-inertial sensor types (GPS, mono camera + IMU, stereo cameras + IMU, even stereo cameras only). It has online spatial calibration, online temporal calibration and visual loop closure.
    \item \textbf{ICE-BA}: it (tightly-coupled, optimization-based) presents an incremental, consistent and efficient bundle adjustment for visual-inertial SLAM, which  performs in parallel both local BA over the sliding window and global BA over all keyframes, and outputs camera pose and updated map points for each frame in real-time \cite{liu2018ice}.
    \item \textbf{Maplab}: it (tightly-coupled, optimization-based) is an open, research-oriented visual-inertial mapping framework, written in C++, for creating, processing and manipulating multi-session maps. On the one hand, maplab can be considered as a ready-to-use visual-inertial mapping and localization system. On the other hand, maplab provides the research community with a collection of multi-session mapping tools that include map merging, visual-inertial batch optimization, loop closure, 3D dense reconstruction \cite{schneider2018maplab}.
\end{itemize}

Other solutions can be listed as follows but not limited to \textbf{VI-ORB} (tightly-coupled, optimization-based) \cite{mur2017visual} (the works by the author of ORB-SLAM, but not open source), \textbf{StructVIO} \cite{zou2019structvio}. \textbf{RKSLAM} \cite{liu2016robust} can reliably handle fast motion and strong rotation for AR applications. Other VIO system based on event camera can be listed as follows but not limited to \cite{mueggler2018continuous}\cite{zhu2017event}\cite{nelson2019event}. \textbf{mi-VINS} \cite{Eckenhoff2019SensorFailureResilientMV} uses multiple IMU, which can work if IMU sensor failures.

VIO SLAM based on deep learning can be seen in \cite{shamwell2019unsupervised}. It shows a network that performs visual-inertial odometry (VIO) without inertial measurement unit (IMU) intrinsic parameters or the extrinsic calibration between an IMU and camera. \cite{lee2019visual} provides a network to avoid the calibration between camera and IMU. 

\subsubsection{Deep Learning with Visual SLAM}
Nowadays, deep learning plays a critical role in the maintenance of computer vision. As the development of visual SLAM, more and more focus are paid into deep learning with SLAM. The term "semantic SLAM" refers to an approach that includes the semantic information into the SLAM process to enhance the performance and representation by providing high-level understanding, robust performance, resource awareness, and task driven perception. Next, we will introduce the implement of SLAM with semantic information in these aspects: 

\begin{itemize}
    \item \textbf{Feature \& Detection}: \textbf{Pop-up SLAM} (Monocular) \cite{yang2016pop} proposes real-time monocular plane SLAM to demonstrate that scene understanding could improve both state estimation and dense mapping especially in low-texture environments. The plane measurements come from a pop-up 3D plane model applied to each single image. \cite{pavlakos20176} gets semantic key points predicted by a convolutional network (convnet). \textbf{LIFT} \cite{yi2016lift} can get more dense feature points than SIFT. \textbf{DeepSLAM} \cite{detone2017toward} has a significant performance gap in the presence of image noise when catch the feature points. \textbf{SuperPoint} \cite{detone2018superpoint} presents a self-supervised framework for training interest point detectors and descriptors suitable for a large number of multiple-view geometry problems in computer vision. \cite{Li2018Stereo} proposes to use the easy-to-labeled 2D detection and discrete viewpoint classification together with a light-weight semantic inference method to obtain rough 3D object measurements. \textbf{GCN-SLAM} \cite{tang2019gcnv2} presents a deep learning-based network, GCNv2, for generation of key points and descriptors. \cite{grinvald2019volumetric} fuses information about 3D shape, location, and, if available, semantic class. \textbf{SalientDSO} \cite{liang2019salientdso} can realize visual saliency and environment perception with the aid of deep learning.  \cite{hosseinzadeh2018structure} integrates the detected objects as the quadrics models into the SLAM system. \textbf{CubeSLAM} (Monocular) is a 3D Object Detection and SLAM system \cite{yang2019cubeslam} based on cube model. It achieve object-level mapping, positioning, and dynamic object tracking. \cite{yang2019monocular} combines the cubeSLAM (high-level object) and Pop-up SLAM (plane landmarks) to make map more denser, more compact and semantic meaningful compared to feature point based SLAM.  \textbf{MonoGRNet} \cite{qin2019monogrnet}  is a geometric reasoning network for monocular 3D object detection and localization. Feature based on event camera can be seen but not limited to \cite{Lagorce2013Event}\cite{mueggler2017fast}. About the survey in deep learning for detection, \cite{wu2019recent} could be a good choice. 
    \item \textbf{Recognition \& Segmentation}: \textbf{SLAM++} (CAD model) \cite{Salas2013SLAM} presents the major advantages of a new ‘object oriented’ 3D SLAM paradigm, which takes full advantage in the loop of prior knowledge that many scenes consist of repeated, domain-specific objects and structures. \cite{Li2016Semi} combines the state-of-art deep learning method and LSD-SLAM based on video stream from a monocular camera. 2D semantic information are transferred to 3D mapping via correspondence between connective keyframes with spatial consistency. \textbf{Semanticfusion} (RGBD) \cite{mccormac2017semanticfusion} combines CNN (Convolutional Neural Network) and a state-of-the-art dense Simultaneous Localization and Mapping (SLAM) system, ElasticFusion \cite{whelan2016elasticfusion} to build a semantic 3D map. \cite{sunderhauf2017meaningful} leverages sparse, feature-based RGB-D SLAM, image-based deep-learning object detection and 3D unsupervised segmentation. \textbf{MarrNet} \cite{wu2017marrnet} proposes an end-to-end trainable framework, sequentially estimating 2.5D sketches and 3D object shapes. \textbf{3DMV} (RGB-D) \cite{dai20183dmv} jointly combines RGB color and geometric information to perform 3D semantic segmentation of RGB-D scans. \textbf{Pix3D} \cite{pix3d} study 3D shape modeling from a single image. \textbf{ScanComplete} \cite{dai2018scancomplete} is a data-driven approach which takes an incomplete 3D scan of a scene as input and predicts a complete 3D model, along with per-voxel semantic labels. \textbf{Fusion++} \cite{McCormac2018FusionVO} is an online object-level SLAM system which builds a persistent and accurate 3D graph map of arbitrary reconstructed objects. As an RGB-D camera browses a cluttered indoor scene, \textbf{Mask-RCNN} instance segmentations are used to initialise compact per-object Truncated Signed Distance Function (TSDF) reconstructions with object size dependent resolutions and a novel 3D foreground mask. \textbf{SegMap} \cite{dube2018segmap} is a map representation based on 3D segments allowing for robot localization, environment reconstruction, and semantics extraction. \textbf{3D-SIS} \cite{hou20193d} is a novel neural network architecture for 3D semantic instance segmentation in commodity RGB-D scans. \textbf{DA-RNN} \cite{xiang2017rnn} uses a new recurrent neural network architecture for semantic labeling on RGB-D videos. \textbf{DenseFusion} \cite{wang2019densefusion} is a generic framework for estimating 6D pose of a set of known objects from RGB-D images. Other work can be seen in \textbf{CCNet} \cite{huang2018ccnet}. To recognize based on event camera, \cite{stromatias2017event}\cite{maro2018event}\cite{afshar2018investigation}\cite{linares2019dynamic} are the best paper to be investigated.
    \item \textbf{Recovery Scale}: \textbf{CNN-SLAM} (Monocular) \cite{tateno2017cnn} estimates the depth with deep learning. Another work can be seen in \textbf{DeepVO} \cite{mohanty2016deepvo}, \textbf{GS3D} \cite{li2019gs3d} . \textbf{UnDeepVO} \cite{li2018undeepvo} can get the 6-DoF pose and the depth using a monocular camera with deep learning. Google proposes the work \cite{li2019learning} that present a method for predicting dense depth in scenarios where both a monocular camera and people in the scene are freely moving based on unsupervised learning. Other methods to get real scale in Monocular can be seen in \cite{8353862}\cite{Sucar2017Bayesian}. \textbf{GeoNet} \cite{yin2018geonet} is a jointly unsupervised learning framework for monocular depth, optical flow and ego-motion estimation from videos. \textbf{CodeSLAM} \cite{bloesch2018codeslam} proposes a depth map from single image, which can be optimised efficiently jointly with pose variables. \textbf{Mono-stixels} \cite{brickwedde2018mono} uses the depth, motion and semantic information in dynamic scene to estimate depth. \textbf{GANVO} \cite{Almalioglu2018GANVOUD} uses an  unsupervised learning framework for 6-DoF pose and monocular depth map from unlabelled image, using deep convolutional Generative Adversarial Networks. \textbf{GEN-SLAM} \cite{chakravarty2019gen} outputs the dense map with the aid of conventional geometric SLAM and the topological constraint in monocular. \cite{Lasinger2019} proposes a training objective that is invariant to changes in depth range and scale. Other similar work can be seen in \textbf{DeepMVS} \cite{huang2018deepmvs} and \textbf{DeepV2D} \cite{DeepV2D}. Based on event camera, depth estimation can be applied in monocular camera \cite{haessig2019spiking}\cite{gallego2018unifying} and stereo camera \cite{xie2017event}. 
    \item \textbf{Pose Output \& Optimization}:  \cite{konda2015learning} is a stereo-VO under the synchronicity. \cite{costante2015exploring} utilizes a CNN to estimate motion from optical flow. \textbf{PoseNet} \cite{kendall2015posenet} can get the 6-DOF pose from a single RGB image without the help of optimization. \textbf{VInet} (Monocular) \cite{clark2017vinet} firstly estimates the motion in VIO, reducing the dependence of manual synchronization and calibration. \textbf{DeepVO} (Monocular) \cite{wang2017deepvo} presents a novel end-to-end framework for monocular VO by using deep Recurrent Convolutional Neural Networks (RCNNs). The similar work can be seen in \textbf{SFMlearner} \cite{zhou2017unsupervised} and \textbf{SFM-Net}\cite{vijayanarasimhan2017sfm}. \textbf{VSO} \cite{Lianos2018VSO} proposes a novel visual semantic odometry (VSO) framework to enable medium-term continuous tracking of points using semantics. \textbf{MID-Fusion} (RGBD, dense point cloud) \cite{xu2018mid} estimates the pose of each existing moving object using an object-oriented tracking method and associate segmented masks with existing models and incrementally fuse corresponding color, depth, semantic, and foreground object probabilities into each object model. Other similar works can be seen in \textbf{VidLoc} \cite{clark2017vidloc}. Besides, \cite{gallego2015event}\cite{reverter2016neuromorphic} are using event camera to output the ego-motion.
    \item \textbf{Long-term Localization}: \cite{Bowman2017Probabilistic} formulates an optimization problem over sensor states and semantic landmark positions that integrates metric information, semantic information, and data associations. \cite{merrill2018lightweight} proposes a novel unsupervised deep neural network architecture of a feature embedding for visual loop closure. \cite{Stenborg2018Long} shows the semantic information is more effective than the traditional feature descriptors. \textbf{X-View} \cite{gawel2018x} leverages semantic graph descriptor matching for global localization, enabling localization under drastically different view-points. \cite{doherty2019multimodal} proposes a solution that represents hypotheses as multiple modes of an equivalent non-Gaussian sensor model to determine object class labels and measurement-landmark correspondences. About the application based on event camera, \cite{censi2013low} are worthy to be read. 
    \item \textbf{Dynamic SLAM}: \textbf{RDSLAM} \cite{tan2013robust}  is a novel real-time monocular SLAM system which can robustly work in dynamic environments based on a novel online keyframe representation and updating method. \textbf{DS-SLAM} \cite{yu2018ds} is a SLAM system with semantic information based on optimized ORB-SLAM. The semantic information can make SLAM system more robust in dynamic environment. \textbf{MaskFusion} (RGB-D, dense point cloud) is a real-time, object-aware, semantic and dynamic RGB-D SLAM system \cite{8613746} based on Mask R-CNN\cite{matterport_maskrcnn_2017}. The system can label the objects with semantic information even in continuously and independent motion. The related work can be seen in \textbf{Co-Fusion} (RGBD)\cite{runz2017co}.  \textbf{Detect-SLAM} \cite{Zhong2018Detect} integrates SLAM with a deep neural network based object detector to make the two functions mutually beneficial in an unknown and dynamic environment. \textbf{DynaSLAM} \cite{bescos2018dynaslam} is a SLAM system for monocular, stereo and RGB-D camera in dynamic environments with aid of static map. \textbf{StaticFusion} \cite{scona2018staticfusion} proposes a method for robust dense RGB-D SLAM in dynamic environments which detects moving objects and simultaneously reconstructs the background structure. The related work based on dynamic environment can be also seen in \textbf{RGB-D SLAM}\cite{dai2018rgb} and \cite{wang2019computationally}\cite{xiao2019dynamic}\cite{barsan2018robust}.
\end{itemize}

Recently, some works utilizes deep-learning to dominate the whole process of SLAM. \textbf{SimVODIS} \cite{kim2019simvodis} can output the depth and the relative pose between frames, while detecting objects and segmenting the object boundaries.

\subsection{Challenge and Future}
\subsubsection{Robustness and Portability}
Visual SLAM still face some important obstacles like the illumination condition, high dynamic environment, fast motion, vigorous rotation and low texture environment. Firstly, global shutter instead of rolling shutter is fundamental to achieve accurate camera pose estimation. Event camera such as dynamic vision sensors is capable of producing up to one million events per second which is enough for very fast motions in high speed and high dynamic range. Secondly, using semantic features like edge, plane, surface features, even reducing feature dependencies, such as tracking with join edges, direct tracking, or a combination of machine learning may become the better choice. Thirdly, based mathematical machinery for SfM/SLAM, the precise mathematical formulations to outperform implicitly learned navigation functions over data is preferred.

The future of SLAM has can be expected that one is SLAM based on smart phones or embedded platforms such as UAV (unmanned aerial vehicle) and another is detailed 3D reconstruction, scene understanding with deep learning.  How to balance real-time and accuracy is the vital open question. The solutions pertaining to dynamic, unstructured, complex, uncertain and large-scale environments are yet to be explored \cite{sualeh2019simultaneous}.
\subsubsection{Multiple Sensors Fusion}
The actual robots and hardware devices usually do not carry only one kind of sensor, and often a fusion of multiple sensors. For example, the current research on VIO on mobile phones combines visual information and IMU information to realize the complementary advantages of the two sensors, which provides a very effective solution for the miniaturization and low cost of SLAM. \textbf{DeLS-3D} \cite{wang2018dels} design is a sensor fusion scheme which integrates camera videos, motion sensors (GPS/IMU), and a 3D semantic map in order to achieve robustness and efficiency of the system. There are sensors listed as follows but not limited to Lidar, Sonar, IMU, IR, camera, GPS, radar, etc. The choice of sensors is dependent on the environment and required type of map.

\subsubsection{Semantics SLAM}
In fact, humans recognize the movement of objects based on perception not the features in image.  Deep learning in SLAM can realize object recognition and segmentation, which help the SLAM system perceive the surrounding better. Semantics SLAM can also do a favor in global optimization, loop closure and relocalization. \cite{atanasov2018unifying}: Traditional approaches for simultaneous localization and mapping (SLAM) depend on geometric features such as points, lines (\textbf{PL-SLAM} \cite{gomez2019pl}, \textbf{StructSLAM} \cite{zhou2015structslam} ), and planes to infer the environment structure. The aim of high-precision real-time positioning in large-scale scenarios could be achieved by semantics SLAM, which teaches robots perceive as humans. 
\subsubsection{Software \& hardware}
SLAM is not an algorithm but an integrated, complex technology \cite{riisgaard2005dummies}. It not only depend on software, but also hardware. The future SLAM system will focus in the deep combination of algorithm and sensors. Based on illustration above, the domain specific processors rather than general processor, integrated sensors module rather than separate sensor like just camera will show great potential. The above work make the developer focus on the algorithm and accelerate the release of real products.

\section{Lidar and Visual SLAM System}
\subsection{Multiple Sensors Calibration}

\begin{itemize}
    \item \textbf{Camera \& IMU}: \textbf{Kalibr} \cite{rehder2016extending} is a toolbox that solves the following calibration problems: Multiple camera calibration, Visual-inertial calibration (camera-IMU) and Rolling Shutter Camera calibration. \textbf{Vins-Fusion} \cite{qin2018online} has online spatial calibration  and online temporal calibration.  \textbf{MSCKF-VIO} \cite{sun2018robust} also has the calibration for camera and IMU. \textbf{mc-VINS} \cite{Eckenhoff2019MultiCameraVN} can calibrate the extrinsic parameters and time offset between all multiple cameras and IMU. Besides, \textbf{IMU-TK} \cite{tpm_icra2014}\cite{pg_imeko2014} can calibrate internal parameter of IMU. Other work can be seen in \cite{li2014high}. \cite{chen2019selective} proposes a end to end network for monocular VIO, which fuses data from camera and IMU.
    \item \textbf{Camera \& Depth}: \textbf{BAD SLAM} \cite{Schops_2019_CVPR} proposes a calibrated benchmark for this task that uses synchronized global shutter RGB and depth cameras.
    \item \textbf{Camera \& Camera}: \textbf{mcptam} \cite{harmat2015multi} is a SLAM system using multi-camera. It can also calibrate the intrinsic and extrinsic parameters. \textbf{MultiCol-SLAM} \cite{UrbanMultiColSLAM16} is a multi-fisheye camera SLAM. Besides, the updated version of SVO can also support multiple cameras. Other similar work can be seen in \textbf{ROVIO} \cite{bloesch2017iterated}.
    \item \textbf{Lidar \& IMU}: \textbf{LIO-mapping} \cite{ye2019tightly} introduces a tightly coupled lidar-IMU fusion method. \textbf{Lidar-Align} is a simple method for finding the extrinsic calibration between a 3D Lidar and a 6-Dof pose sensor. Extrinsic calibration of Lidar can be seen in \cite{yin2018extrinsic}\cite{chen2018extrinsic}. The doctoral thesis \cite{levinson2011automatic} illustrate the work of Lidar calibration. 
    \item \textbf{Camera \& Lidar}: \cite{levinson2013automatic} introduces a probabilistic monitoring algorithm and a continuous calibration optimizer that enable camera-laser calibration online, automatically. \textbf{Lidar-Camera} \cite{2017arXiv170509785D} proposes a novel pipeline and experimental setup to find accurate rigid-body transformation for extrinsically calibrating a LiDAR and a camera using 3D-3D point correspondences. \textbf{RegNet} \cite{schneider2017regnet} is the first deep convolutional neural network (CNN) to infer a 6 degrees of freedom (DOF) extrinsic calibration between multi-modal sensors, exemplified using a scanning LiDAR and a monocular camera. \textbf{LIMO} \cite{Graeter2018LIMO} proposes a depth extraction algorithm from LIDAR measurements for camera feature tracks and estimating motion. \textbf{CalibNet} \cite{iyer2018calibnet} is  a self-supervised deep network capable of automatically estimating the 6-DoF rigid body transformation between a 3D LiDAR and a 2D camera in real-time. The calibration tool from \textbf{Autoware} can calibrate the signal beam Lidar and camera. . Other work can be seen as follows but not limited to \cite{mirzaei20123d}\cite{ishikawa2018lidar}\cite{Levinson2013AutomaticOC}.

\end{itemize}

Other work like \textbf{SVIn2} \cite{rahmansvin2} demonstrates an underwater SLAM system fusing Sonar, Visual, Inertial, and Depth Sensor, which is based on OKVIS. \cite{Gu2019EnvironmentDU} proposes a new underwater camera-IMU calibration model and \cite{Arain2019ImprovingUO} detects underwater obstacle using semantic image Segmentation. \textbf{WiFi-SLAM} \cite{ferris2007wifi} demonstrates a novel SLAM technology with wireless signal named WiFi. \cite{aladsani2019leveraging} uses the mmWave to locate even the NLOS robots and \cite{kanhere2018position}\cite{rappaport2019wireless} introduce more technique about localization with the aid of wirless signals. \textbf{KO-Fusion} \cite{houseago2019ko} fuses visual and wheeled odometer. \cite{Khattak2019KeyframebasedDT} uses a thermal camera with IMU in visually degraded environments e.g. darkness. 

\subsection{Lidar and Visual Fusion}
\begin{itemize}
    \item \textbf{Hardware layer}: \textbf{Pandora} from HESAI is a software and hardware solution integrating 40 beams Lidar, five color cameras and recognition algorithm. The integrated solution can comfort developer from temporal and spatial synchronization. Understanding the exist of \textbf{CONTOUR} and \textbf{STENCIL} from \textbf{KAARTA} will give you a brainstorming.
    \item \textbf{Data layer}: Lidar has sparse, high precision depth data and camera has dense but low precision depth data, which will lead to image-based depth upsampling and image-based depth inpainting/completion. \cite{ferstl2013image} presents a novel method for the challenging problem of depth image upsampling. \cite{ku2018defense} relies only on basic image processing operations to perform depth completion of sparse Lidar depth data. With deep learning, \cite{mal2018sparse} proposes the use of a single deep regression network to learn directly from the RGB-D raw data, and explore the impact of number of depth samples. \cite{uhrig2017sparsity} considers CNN operating on sparse inputs with an application to depth completion from sparse laser scan data. \textbf{DFuseNet} \cite{shivakumar2019dfusenet} proposes a CNN that is designed to upsample a series of sparse range measurements based on the contextual cues gleaned from a high resolution intensity image. Other similar work can be seen as follows but not limited to \cite{chen2018estimating} \cite{eldesokey2018propagating}. \textbf{LIC-Fusion} \cite{zuo2019lic} fuses IMU measurements, sparse visual features, and extracted LiDAR points.
    \item \textbf{Task layer}: \cite{aycard2011intersection} fuses stereo camera and Lidar to perceive. \cite{chavez2015multiple} fuses radar, Lidar, and camera to detect and classify moving objects. Other traditional work can be seen but not limited to \cite{cho2014multi} \cite{wang2011integrating}\cite{wan2018robust}. \cite{zhang2014real} can augment VO by depth information such as provided by RGB-D cameras, or from Lidars associated with cameras even if sparsely available. \textbf{V-Loam} \cite{zhang2015visual} presents a general framework for combining visual odometry and Lidar odometry. The online method starts with visual odometry and scan matching based Lidar odometry refines the motion estimation and point cloud registration simultaneously. \textbf{VI-SLAM} \cite{nava2018visual} is concerned with the development of a system that combines an accurate laser odometry estimator, with algorithms for place recognition using vision for achieving loop detection. \cite{xu2018slam} aims at the tracking part of SLAM using an RGB-D camera and 2d low-cost LIDAR to finish a robust indoor SLAM by a mode switch and data fusion. \textbf{VIL-SLAM} \cite{shao2019stereo} incorporates tightly-coupled stereo VIO with Lidar mapping and Lidar enhanced visual loop closure. \cite{andert2015lidar} combines monocular camera images with laser distance measurements to allow visual SLAM without errors from increasing scale uncertainty. In deep learning, many methods to detect and recognize fusing data from camera and Lidar such as \textbf{PointFusion} \cite{xu2018pointfusion}, \textbf{RoarNet} \cite{shin2018roarnet}, \textbf{AVOD} \cite{ku2018joint}, \textbf{MV3D} \cite{chen2017multi}, \textbf{FuseNet} \cite{hazirbas2016fusenet}. Other similar work can be seen in \cite{wang2018fusing}. Besides, \cite{liang2018deep} exploits both Lidar as well as cameras to perform very accurate localization with a  an end-to-end learnable architecture. \cite{gu20183} fuses 3D Lidar and monocular camera.

\end{itemize}

\subsection{Challenge and Future \cite{cadena2016past}}

\begin{itemize}
    \item \textbf{Data Association}:  the future of SLAM must integrate multi-sensors. But different sensors have different data types, time stamps, and coordinate system expressions, needed to be processed uniformly. Besides, physical model establishment, state estimation and optimization between multi-sensors should be taken into consideration.
    \item \textbf{Integrated Hardware}: at present, there is no suitable chip and integrated hardware to make technology of SLAM more easily to be a product. On the other hand,  if the accuracy of a sensor degrades due to malfunctioning, off-nominal conditions, or aging, the quality of the sensor measurements (e.g., noise, bias) does not match the noise model. The robustness and integration of hardware should be followed. Sensors in front-end should have the capability to process data and the evolution from hardware layer to algorithm layer, then to function layer to SDK should be innovated to application. 
    \item \textbf{Crowdsourcing}: decentralized visual SLAM is a powerful tool for multi-robot applications in environments where absolute positioning systems are not available \cite{cieslewski2018data}. Co-optimization visual multi-robot SLAM need decentralized data and optimization, which is called crowdsourcing. The privacy in the process of decentralized data should come into attention. The technology of differential privacy  \cite{dwork2011differential}\cite{mcsherry2007mechanism}  maybe do a favor. 
    \item \textbf{High Definition Map}: High Definition Map is vital for robots. But which type of map is the best for robots? Could dense map or sparse map navigate, positioning and path plan? A related open question for long-term mapping is how often to update the information contained in the map and how to decide when this information becomes outdated and can be discarded.
    \item \textbf{Adaptability, Robustness, Scalability}:  as we know, no SLAM system now can cover all scenarios. Most of it requires extensive parameter tuning in order to work correctly for a given scenario. To make robots perceive as humans, appearance-based instead of feature-based method is preferred, which will help close loops integrated with semantic information between day and night sequences or between different seasons. 
    \item \textbf{Ability against risk and constraints}: Perfect SLAM system should be failure-safe and failure-aware. It's not the question about relocalization or loop closure here. SLAM system must have ability to response to risk or failure. In the same time, an ideal SLAM solution should be able run on different platforms no matter the computational constraints of platforms. How to balance the accuracy, robustness and the limited resource is a challenging problem \cite{delmerico2018benchmark}. 
    \item \textbf{Application}: the technology of SLAM has a wide application such as: large-scale positioning, navigation and 3D or semantic map construction, environment recognition and understanding,  ground robotics, UAV, VR/AR/MR, AGV(Automatic Guided Vehicle), automatic drive, virtual interior decorator, virtual fitting room, immersive online game, earthquake relief, video segmentation and editing
    \item \textbf{Open question}: Will end-to-end learning dominate SLAM? 
\end{itemize}

\section{An Envision in 6G Wireless Networks}
Nowadays, 5G has been developed widely to communicate more quickly and massively \cite{rappaport2019wireless}. But for robots and autonomous driving cars, the technology of SLAM need greater data rates and less latency that 5G can't afford. Unlike 100 Gbps of data rates for 5G, 6G can provide greater data rates due to the frequency in 100 GHz to 3 THz (terahertz). 

THz is the last unexplored band in the radio frequency spectrum. Less than the THz, the radio bands are called microwave. The radio frequency of optical bands , which are regarded as visible light communications (VLC), are more than THz \cite{sarieddeen2019next}. The technology of 6G will need no supports such as multiple-input multiple-output (MIMO) in 5G represented as mmWave communications. As for the difference with VLC, 6G with the THz communications will not affected by the light changes and NLOS.

\subsection{The advantages of  6G Wireless Networks}
For the advance of wireless communication system, industry and academic are urged to pay attention to the research of 6G. Next, we will introduce some advantages of 6G \cite{letaief2019roadmap}.  
\begin{itemize}
	\item \textbf{Low Latency}: Less than 1 msec end-to-end latency; 
	\item \textbf{Data rate}: High data rates up to 1 Tbps;
	\item \textbf{Ultra-high bandwidth}: Very broad frequency bands;
	\item \textbf{Energy save}: Very high energy efficiency; 
	\item \textbf{Ubiquitous connection}: Enable to connect global network including the massive intelligent things and the emergence of smart surface and environment such as walls, roads even the whole buildings. 
	\item \textbf{Intelligent network}: AI and RISs make it smarter and beyond classical big data analytics and edge computation.
\end{itemize}
6G is supposed as the platforms to serve for communication, computation, and storage resources with the aid of AI \cite{saad2019vision}. It make the society fully connected and digital.

\subsection{6G in Simultaneous Localization and Mapping}

SLAM can be divided into radio-based (such as satellite positioning, cellular and WiFi) and sensor-based (such as Lidar, IMU and camera) \cite{wymeersch2019radio}. With the wireless technology, the technology of SLAM can be achieved by constructing the map of environment and projecting the angle and the time of arrival to estimate the locations from the users. The conventional mmWave method utilizing  AoA-based positioning or a combination of path loss and AoA, and RSSI \cite{kanhere2018position}. With the aid of Reconfigurable intelligent surfaces (RISs),  localization and mapping can improve accuracy and extended physical coverage.

It's apparent that 6G with THz will create centimeter level accuracy even in NLOS environment and 6G will provide a network to sense and localization rather than a independent source \cite{giordani2019towards} for SLAM. Cause the greater data rates, the massive computations can be conducted in remote device or machine, which relief the pressure of computer power in robots and autonomous driving cars. 

In the future, THz will enable the robots and autonomous driving cars with new capability of sensing the gas, air quality, health detection, body scanning and so on. Plus, with the aid of THz, computer vision will be augmented to see the NLOS views, which will play a vital role in rescue and sensing. In the basic, the accuracy of positioning will be improved to sub-centimeter level and the map of surrounding environment will be constructed as 3D maps without any calibration and prior knowledge, which is hard to realized before. Moreover, 6G and AI will achieve excellent successes again in future digital society with the full connectivity demands.

\section{Conclusion}
Simultaneous Localization and Mapping (SLAM) based on Lidar and camera sensors are introduced in this paper. Started with the type and brand of sensors, the different categories of SLAM present the development with the aid of open-source solutions.

2D Lidar SLAM, 3D Lidar SLAM, deep learning in Lidar and challenge as well as future illustrate the general view over Lidar. Although the stability of the Lidar sensor, visual SLAM based on camera sensors is one of the most focused topics due to the prosperity of deep learning. This paper highlights the visual SLAM including camera sensors, the different density of open-source visual SLAM system, visual-inertial odometry SLAM and deep learning in visual SLAM and future. At present, many kinds of researches pay attention to the fusion of Lidar and vision, which is a balanced way for reliability and versatility. 

For the future, different solutions with SLAM in diverse applications will face challenges and chances. This paper points out several directions for future research of SLAM in sensor-based and provides an envision in 6G development to SLAM in radio-based.

\ifCLASSOPTIONcaptionsoff
  \newpage
\fi

\bibliography{paper}

\begin{thebibliography}{100}

\bibitem{leonard1991simultaneous}
John~J Leonard and Hugh~F Durrant-Whyte.
\newblock Simultaneous map building and localization for an autonomous mobile
  robot.
\newblock In {\em Proceedings IROS'91: IEEE/RSJ International Workshop on
  Intelligent Robots and Systems' 91}, pages 1442--1447. Ieee, 1991.

\bibitem{smith1990estimating}
Randall Smith, Matthew Self, and Peter Cheeseman.
\newblock Estimating uncertain spatial relationships in robotics.
\newblock In {\em Autonomous robot vehicles}, pages 167--193. Springer, 1990.

\bibitem{huang2019robust}
Baichuan Huang, Jingbin Liu, Wei Sun, and Fan Yang.
\newblock A robust indoor positioning method based on bluetooth low energy with
  separate channel information.
\newblock {\em Sensors}, 19(16):3487, 2019.

\bibitem{liu2012iparking}
Jingbin Liu, Ruizhi Chen, Yuwei Chen, Ling Pei, and Liang Chen.
\newblock iparking: An intelligent indoor location-based smartphone parking
  service.
\newblock {\em Sensors}, 12(11):14612--14629, 2012.

\bibitem{liu2012hybrid}
Jingbin Liu, Ruizhi Chen, Ling Pei, Robert Guinness, and Heidi Kuusniemi.
\newblock A hybrid smartphone indoor positioning solution for mobile lbs.
\newblock {\em Sensors}, 12(12):17208--17233, 2012.

\bibitem{thrun2005probabilistic}
Sebastian Thrun, Wolfram Burgard, and Dieter Fox.
\newblock {\em Probabilistic robotics}.
\newblock MIT press, 2005.

\bibitem{santos2013evaluation}
Joao~Machado Santos, David Portugal, and Rui~P Rocha.
\newblock An evaluation of 2d slam techniques available in robot operating
  system.
\newblock In {\em 2013 IEEE International Symposium on Safety, Security, and
  Rescue Robotics (SSRR)}, pages 1--6. IEEE, 2013.

\bibitem{grisetti2007improved}
Giorgio Grisetti, Cyrill Stachniss, Wolfram Burgard, et~al.
\newblock Improved techniques for grid mapping with rao-blackwellized particle
  filters.
\newblock {\em IEEE transactions on Robotics}, 23(1):34, 2007.

\bibitem{montemerlo2002fastslam}
Michael Montemerlo, Sebastian Thrun, Daphne Koller, Ben Wegbreit, et~al.
\newblock Fastslam: A factored solution to the simultaneous localization and
  mapping problem.
\newblock {\em Aaai/iaai}, 593598, 2002.

\bibitem{montemerlo2003fastslam}
Michael Montemerlo, Sebastian Thrun, Daphne Koller, Ben Wegbreit, et~al.
\newblock Fastslam 2.0: An improved particle filtering algorithm for
  simultaneous localization and mapping that provably converges.
\newblock In {\em IJCAI}, pages 1151--1156, 2003.

\bibitem{kohlbrecher2011flexible}
Stefan Kohlbrecher, Oskar Von~Stryk, Johannes Meyer, and Uwe Klingauf.
\newblock A flexible and scalable slam system with full 3d motion estimation.
\newblock In {\em 2011 IEEE International Symposium on Safety, Security, and
  Rescue Robotics}, pages 155--160. IEEE, 2011.

\bibitem{konolige2010efficient}
Kurt Konolige, Giorgio Grisetti, Rainer K{\"u}mmerle, Wolfram Burgard, Benson
  Limketkai, and Regis Vincent.
\newblock Efficient sparse pose adjustment for 2d mapping.
\newblock In {\em 2010 IEEE/RSJ International Conference on Intelligent Robots
  and Systems}, pages 22--29. IEEE, 2010.

\bibitem{carlone2012linear}
Luca Carlone, Rosario Aragues, Jos{\'e}~A Castellanos, and Basilio Bona.
\newblock A linear approximation for graph-based simultaneous localization and
  mapping.
\newblock {\em Robotics: Science and Systems VII}, pages 41--48, 2012.

\bibitem{steux2010slam}
B~Steux and O~TinySLAM El~Hamzaoui.
\newblock A slam algorithm in less than 200 lines c-language program.
\newblock {\em Proceedings of the Control Automation Robotics \& Vision
  (ICARCV), Singapore}, pages 7--10, 2010.

\bibitem{hess2016real}
Wolfgang Hess, Damon Kohler, Holger Rapp, and Daniel Andor.
\newblock Real-time loop closure in 2d lidar slam.
\newblock In {\em 2016 IEEE International Conference on Robotics and Automation
  (ICRA)}, pages 1271--1278. IEEE, 2016.

\bibitem{zhang2014loam}
Ji~Zhang and Sanjiv Singh.
\newblock Loam: Lidar odometry and mapping in real-time.
\newblock In {\em Robotics: Science and Systems}, volume~2, page~9, 2014.

\bibitem{shan2018lego}
Tixiao Shan and Brendan Englot.
\newblock Lego-loam: Lightweight and ground-optimized lidar odometry and
  mapping on variable terrain.
\newblock In {\em 2018 IEEE/RSJ International Conference on Intelligent Robots
  and Systems (IROS)}, pages 4758--4765. IEEE, 2018.

\bibitem{deschaud2018imls}
Jean-Emmanuel Deschaud.
\newblock Imls-slam: scan-to-model matching based on 3d data.
\newblock In {\em 2018 IEEE International Conference on Robotics and Automation
  (ICRA)}, pages 2480--2485. IEEE, 2018.

\bibitem{angelina2018pointnetvlad}
Mikaela Angelina~Uy and Gim Hee~Lee.
\newblock Pointnetvlad: Deep point cloud based retrieval for large-scale place
  recognition.
\newblock In {\em Proceedings of the IEEE Conference on Computer Vision and
  Pattern Recognition}, pages 4470--4479, 2018.

\bibitem{zhou2018voxelnet}
Yin Zhou and Oncel Tuzel.
\newblock Voxelnet: End-to-end learning for point cloud based 3d object
  detection.
\newblock In {\em Proceedings of the IEEE Conference on Computer Vision and
  Pattern Recognition}, pages 4490--4499, 2018.

\bibitem{beltran2018birdnet}
Jorge Beltr{\'a}n, Carlos Guindel, Francisco~Miguel Moreno, Daniel Cruzado,
  Fernando Garcia, and Arturo De~La~Escalera.
\newblock Birdnet: a 3d object detection framework from lidar information.
\newblock In {\em 2018 21st International Conference on Intelligent
  Transportation Systems (ITSC)}, pages 3517--3523. IEEE, 2018.

\bibitem{minemura2018lmnet}
Kazuki Minemura, Hengfui Liau, Abraham Monrroy, and Shinpei Kato.
\newblock Lmnet: Real-time multiclass object detection on cpu using 3d lidar.
\newblock In {\em 2018 3rd Asia-Pacific Conference on Intelligent Robot Systems
  (ACIRS)}, pages 28--34. IEEE, 2018.

\bibitem{yang2018pixor}
Bin Yang, Wenjie Luo, and Raquel Urtasun.
\newblock Pixor: Real-time 3d object detection from point clouds.
\newblock In {\em Proceedings of the IEEE conference on Computer Vision and
  Pattern Recognition}, pages 7652--7660, 2018.

\bibitem{ali2018yolo3d}
Waleed Ali, Sherif Abdelkarim, Mahmoud Zidan, Mohamed Zahran, and Ahmad
  El~Sallab.
\newblock Yolo3d: End-to-end real-time 3d oriented object bounding box
  detection from lidar point cloud.
\newblock In {\em Proceedings of the European Conference on Computer Vision
  (ECCV)}, pages 0--0, 2018.

\bibitem{li2018pointcnn}
Yangyan Li, Rui Bu, Mingchao Sun, Wei Wu, Xinhan Di, and Baoquan Chen.
\newblock Pointcnn: Convolution on x-transformed points.
\newblock In {\em Advances in Neural Information Processing Systems}, pages
  820--830, 2018.

\bibitem{chen2017multi}
Xiaozhi Chen, Huimin Ma, Ji~Wan, Bo~Li, and Tian Xia.
\newblock Multi-view 3d object detection network for autonomous driving.
\newblock In {\em Proceedings of the IEEE Conference on Computer Vision and
  Pattern Recognition}, pages 1907--1915, 2017.

\bibitem{li2019pu}
Ruihui Li, Xianzhi Li, Chi-Wing Fu, Daniel Cohen-Or, and Pheng-Ann Heng.
\newblock Pu-gan: A point cloud upsampling adversarial network.
\newblock In {\em Proceedings of the IEEE International Conference on Computer
  Vision}, pages 7203--7212, 2019.

\bibitem{su2018splatnet}
Hang Su, Varun Jampani, Deqing Sun, Subhransu Maji, Evangelos Kalogerakis,
  Ming-Hsuan Yang, and Jan Kautz.
\newblock Splatnet: Sparse lattice networks for point cloud processing.
\newblock In {\em Proceedings of the IEEE Conference on Computer Vision and
  Pattern Recognition}, pages 2530--2539, 2018.

\bibitem{grilli2017review}
E~Grilli, F~Menna, and F~Remondino.
\newblock A review of point clouds segmentation and classification algorithms.
\newblock {\em The International Archives of Photogrammetry, Remote Sensing and
  Spatial Information Sciences}, 42:339, 2017.

\bibitem{qi2016pointnet}
Charles~R Qi, Hao Su, Kaichun Mo, and Leonidas~J Guibas.
\newblock Pointnet: Deep learning on point sets for 3d classification and
  segmentation.
\newblock {\em arXiv preprint arXiv:1612.00593}, 2016.

\bibitem{qi2017pointnetplusplus}
Charles~R Qi, Li~Yi, Hao Su, and Leonidas~J Guibas.
\newblock Pointnet++: Deep hierarchical feature learning on point sets in a
  metric space.
\newblock {\em arXiv preprint arXiv:1706.02413}, 2017.

\bibitem{qi2019deep}
Charles~R Qi, Or~Litany, Kaiming He, and Leonidas~J Guibas.
\newblock Deep hough voting for 3d object detection in point clouds.
\newblock {\em arXiv preprint arXiv:1904.09664}, 2019.

\bibitem{segmap2018}
Renaud Dube, Andrei Cramariuc, Daniel Dugas, Juan Nieto, Roland Siegwart, and
  Cesar Cadena.
\newblock {SegMap}: 3d segment mapping using data-driven descriptors.
\newblock In {\em Robotics: Science and Systems (RSS)}, 2018.

\bibitem{wu2017squeezeseg}
Bichen Wu, Alvin Wan, Xiangyu Yue, and Kurt Keutzer.
\newblock Squeezeseg: Convolutional neural nets with recurrent crf for
  real-time road-object segmentation from 3d lidar point cloud.
\newblock {\em ICRA}, 2018.

\bibitem{wu2018squeezesegv2}
Bichen Wu, Xuanyu Zhou, Sicheng Zhao, Xiangyu Yue, and Kurt Keutzer.
\newblock Squeezesegv2: Improved model structure and unsupervised domain
  adaptation for road-object segmentation from a lidar point cloud.
\newblock In {\em ICRA}, 2019.

\bibitem{yue2018lidar}
Xiangyu Yue, Bichen Wu, Sanjit~A Seshia, Kurt Keutzer, and Alberto~L
  Sangiovanni-Vincentelli.
\newblock A lidar point cloud generator: from a virtual world to autonomous
  driving.
\newblock In {\em ICMR}, pages 458--464. ACM, 2018.

\bibitem{jiang2018pointsift}
Mingyang Jiang, Yiran Wu, Tianqi Zhao, Zelin Zhao, and Cewu Lu.
\newblock Pointsift: A sift-like network module for 3d point cloud semantic
  segmentation.
\newblock {\em arXiv preprint arXiv:1807.00652}, 2018.

\bibitem{hua-pointwise-cvpr18}
Binh-Son Hua, Minh-Khoi Tran, and Sai-Kit Yeung.
\newblock Pointwise convolutional neural networks.
\newblock In {\em Computer Vision and Pattern Recognition (CVPR)}, 2018.

\bibitem{ye20183d}
Xiaoqing Ye, Jiamao Li, Hexiao Huang, Liang Du, and Xiaolin Zhang.
\newblock 3d recurrent neural networks with context fusion for point cloud
  semantic segmentation.
\newblock In {\em Proceedings of the European Conference on Computer Vision
  (ECCV)}, pages 403--417, 2018.

\bibitem{landrieu2018large}
Loic Landrieu and Martin Simonovsky.
\newblock Large-scale point cloud semantic segmentation with superpoint graphs.
\newblock In {\em Proceedings of the IEEE Conference on Computer Vision and
  Pattern Recognition}, pages 4558--4567, 2018.

\bibitem{dube2017segmatch}
Renaud Dub{\'e}, Daniel Dugas, Elena Stumm, Juan Nieto, Roland Siegwart, and
  Cesar Cadena.
\newblock Segmatch: Segment based place recognition in 3d point clouds.
\newblock In {\em 2017 IEEE International Conference on Robotics and Automation
  (ICRA)}, pages 5266--5272. IEEE, 2017.

\bibitem{klokov2017escape}
Roman Klokov and Victor Lempitsky.
\newblock Escape from cells: Deep kd-networks for the recognition of 3d point
  cloud models.
\newblock In {\em Proceedings of the IEEE International Conference on Computer
  Vision}, pages 863--872, 2017.

\bibitem{dewan2019deeptemporalseg}
Ayush Dewan and Wolfram Burgard.
\newblock Deeptemporalseg: Temporally consistent semantic segmentation of 3d
  lidar scans.
\newblock {\em arXiv preprint arXiv:1906.06962}, 2019.

\bibitem{LU-Net2019}
Pierre Biasutti, Vincent Lepetit, Jean-François Aujol, Mathieu Brédif, and
  Aurélie Bugeau.
\newblock Lu-net: An efficient network for 3d lidar point cloud semantic
  segmentation based on end-to-end-learned 3d features and u-net.
\newblock 08 2019.

\bibitem{shi2019pointrcnn}
Shaoshuai Shi, Xiaogang Wang, and Hongsheng Li.
\newblock Pointrcnn: 3d object proposal generation and detection from point
  cloud.
\newblock In {\em Proceedings of the IEEE Conference on Computer Vision and
  Pattern Recognition}, pages 770--779, 2019.

\bibitem{L3-Net}
Lu~Weixin, Zhou Yao, Wan Guowei, Hou Shenhua, and Song Shiyu.
\newblock L3-net: Towards learning based lidar localization for autonomous
  driving.
\newblock In {\em IEEE Conference on Computer Vision and Pattern Recognition
  (CVPR)}, 2019.

\bibitem{Chen2019suma}
Chen Xieyuanli, Milioto Andres, and Emanuelea Palazzolo.
\newblock Suma++: Efficient lidar-based semantic slam.
\newblock In {\em 2019 IEEE/RSJ International Conference on Intelligent Robots
  and Systems (IROS)}. IEEE, 2019.

\bibitem{wang2018imu}
Zhongli Wang, Yan Chen, Yue Mei, Kuo Yang, and Baigen Cai.
\newblock Imu-assisted 2d slam method for low-texture and dynamic environments.
\newblock {\em Applied Sciences}, 8(12):2534, 2018.

\bibitem{walcott2012dynamic}
Aisha Walcott-Bryant, Michael Kaess, Hordur Johannsson, and John~J Leonard.
\newblock Dynamic pose graph slam: Long-term mapping in low dynamic
  environments.
\newblock In {\em 2012 IEEE/RSJ International Conference on Intelligent Robots
  and Systems}, pages 1871--1878. IEEE, 2012.

\bibitem{shin2017illusion}
Hocheol Shin, Dohyun Kim, Yujin Kwon, and Yongdae Kim.
\newblock Illusion and dazzle: Adversarial optical channel exploits against
  lidars for automotive applications.
\newblock In {\em International Conference on Cryptographic Hardware and
  Embedded Systems}, pages 445--467. Springer, 2017.

\bibitem{cao2019adversarial}
Yulong Cao, Chaowei Xiao, Benjamin Cyr, Yimeng Zhou, Won Park, Sara Rampazzi,
  Qi~Alfred Chen, Kevin Fu, and Z~Morley Mao.
\newblock Adversarial sensor attack on lidar-based perception in autonomous
  driving.
\newblock {\em arXiv preprint arXiv:1907.06826}, 2019.

\bibitem{mur2015orb}
Raul Mur-Artal, Jose Maria~Martinez Montiel, and Juan~D Tardos.
\newblock Orb-slam: a versatile and accurate monocular slam system.
\newblock {\em IEEE transactions on robotics}, 31(5):1147--1163, 2015.

\bibitem{qin2018vins}
Tong Qin, Peiliang Li, and Shaojie Shen.
\newblock Vins-mono: A robust and versatile monocular visual-inertial state
  estimator.
\newblock {\em IEEE Transactions on Robotics}, 34(4):1004--1020, 2018.

\bibitem{klein2009parallel}
Georg Klein and David Murray.
\newblock Parallel tracking and mapping on a camera phone.
\newblock In {\em 2009 8th IEEE International Symposium on Mixed and Augmented
  Reality}, pages 83--86. IEEE, 2009.

\bibitem{InfiniTAM_ISMAR_2015}
O.~{K{\"a}hler}, V.~A. {Prisacariu}, C.~Y. {Ren}, X.~{Sun}, P.~H.~S {Torr}, and
  D.~W. {Murray}.
\newblock {Very High Frame Rate Volumetric Integration of Depth Images on
  Mobile Device}.
\newblock {\em {IEEE Transactions on Visualization and Computer Graphics
  (Proceedings International Symposium on Mixed and Augmented Reality 2015}},
  22(11), 2015.

\bibitem{lynen2015get}
Simon Lynen, Torsten Sattler, Michael Bosse, Joel~A Hesch, Marc Pollefeys, and
  Roland Siegwart.
\newblock Get out of my lab: Large-scale, real-time visual-inertial
  localization.
\newblock In {\em Robotics: Science and Systems}, volume~1, 2015.

\bibitem{Gao2017SLAM}
Xiang Gao, Tao Zhang, Yi~Liu, and Qinrui Yan.
\newblock {\em 14 Lectures on Visual SLAM: From Theory to Practice}.
\newblock Publishing House of Electronics Industry, 2017.

\bibitem{taketomi2017visual}
Takafumi Taketomi, Hideaki Uchiyama, and Sei Ikeda.
\newblock Visual slam algorithms: A survey from 2010 to 2016.
\newblock {\em IPSJ Transactions on Computer Vision and Applications}, 9(1):16,
  2017.

\bibitem{reddy1996fft}
B~Srinivasa Reddy and Biswanath~N Chatterji.
\newblock An fft-based technique for translation, rotation, and scale-invariant
  image registration.
\newblock {\em IEEE transactions on image processing}, 5(8):1266--1271, 1996.

\bibitem{kazik2011visual}
Tim Kazik and Ali~Haydar G{\"o}kto{\u{g}}an.
\newblock Visual odometry based on the fourier-mellin transform for a rover
  using a monocular ground-facing camera.
\newblock In {\em 2011 IEEE International Conference on Mechatronics}, pages
  469--474. IEEE, 2011.

\bibitem{birem2018visual}
Merwan Birem, Richard Kleihorst, and Norddin El-Ghouti.
\newblock Visual odometry based on the fourier transform using a monocular
  ground-facing camera.
\newblock {\em Journal of Real-Time Image Processing}, 14(3):637--646, 2018.

\bibitem{Zhangguofeng2016}
Liu Haomin, Zhang Guofeng, and Bao hujun.
\newblock A survy of monocular simultaneous localization and mapping.
\newblock {\em Journal of Computer-Aided Design \& Computer Graphics},
  28(6):855--868, 2016.

\bibitem{Gallego2019Event}
Guillermo Gallego, Tobi Delbruck, Garrick Orchard, Chiara Bartolozzi, and
  Davide Scaramuzza.
\newblock Event-based vision: A survey.
\newblock 2019.

\bibitem{lichtsteiner2008128}
Patrick Lichtsteiner, Christoph Posch, and Tobi Delbruck.
\newblock A 128x128 120db 15us latency asynchronous temporal contrast vision
  sensor.
\newblock {\em IEEE journal of solid-state circuits}, 43(2):566--576, 2008.

\bibitem{son20174}
Bongki Son, Yunjae Suh, Sungho Kim, Heejae Jung, Jun-Seok Kim, Changwoo Shin,
  Keunju Park, Kyoobin Lee, Jinman Park, Jooyeon Woo, et~al.
\newblock 4.1 a 640$\times$ 480 dynamic vision sensor with a 9$\mu$m pixel and
  300meps address-event representation.
\newblock In {\em 2017 IEEE International Solid-State Circuits Conference
  (ISSCC)}, pages 66--67. IEEE, 2017.

\bibitem{posch2009microbolometer}
Christoph Posch, Daniel Matolin, Rainer Wohlgenannt, Thomas Maier, and Martin
  Litzenberger.
\newblock A microbolometer asynchronous dynamic vision sensor for lwir.
\newblock {\em IEEE Sensors Journal}, 9(6):654--664, 2009.

\bibitem{hofstatter2010sparc}
Michael Hofstatter, Peter Sch{\"o}n, and Christoph Posch.
\newblock A sparc-compatible general purpose address-event processor with
  20-bit l0ns-resolution asynchronous sensor data interface in 0.18 $\mu$m
  cmos.
\newblock In {\em Proceedings of 2010 IEEE International Symposium on Circuits
  and Systems}, pages 4229--4232. IEEE, 2010.

\bibitem{posch2007dual}
Christoph Posch, Michael Hofstatter, Daniel Matolin, Guy Vanstraelen, Peter
  Schon, Nikolaus Donath, and Martin Litzenberger.
\newblock A dual-line optical transient sensor with on-chip precision
  time-stamp generation.
\newblock In {\em 2007 IEEE International Solid-State Circuits Conference.
  Digest of Technical Papers}, pages 500--618. IEEE, 2007.

\bibitem{brandli2014240}
Christian Brandli, Raphael Berner, Minhao Yang, Shih-Chii Liu, and Tobi
  Delbruck.
\newblock A 240$\times$ 180 130 db 3 $\mu$s latency global shutter
  spatiotemporal vision sensor.
\newblock {\em IEEE Journal of Solid-State Circuits}, 49(10):2333--2341, 2014.

\bibitem{posch2010qvga}
Christoph Posch, Daniel Matolin, and Rainer Wohlgenannt.
\newblock A qvga 143 db dynamic range frame-free pwm image sensor with lossless
  pixel-level video compression and time-domain cds.
\newblock {\em IEEE Journal of Solid-State Circuits}, 46(1):259--275, 2010.

\bibitem{davison2007monoslam}
Andrew~J Davison, Ian~D Reid, Nicholas~D Molton, and Olivier Stasse.
\newblock Monoslam: Real-time single camera slam.
\newblock {\em IEEE Transactions on Pattern Analysis \& Machine Intelligence},
  (6):1052--1067, 2007.

\bibitem{klein2007parallel}
Georg Klein and David Murray.
\newblock Parallel tracking and mapping for small ar workspaces.
\newblock In {\em Proceedings of the 2007 6th IEEE and ACM International
  Symposium on Mixed and Augmented Reality}, pages 1--10. IEEE Computer
  Society, 2007.

\bibitem{klein2008improving}
Georg Klein and David Murray.
\newblock Improving the agility of keyframe-based slam.
\newblock In {\em European Conference on Computer Vision}, pages 802--815.
  Springer, 2008.

\bibitem{rublee2011orb}
Ethan Rublee, Vincent Rabaud, Kurt Konolige, and Gary~R Bradski.
\newblock Orb: An efficient alternative to sift or surf.
\newblock In {\em ICCV}, volume~11, page~2. Citeseer, 2011.

\bibitem{mur2017orb}
Raul Mur-Artal and Juan~D Tard{\'o}s.
\newblock Orb-slam2: An open-source slam system for monocular, stereo, and
  rgb-d cameras.
\newblock {\em IEEE Transactions on Robotics}, 33(5):1255--1262, 2017.

\bibitem{wang2018cubemapslam}
Yahui Wang, Shaojun Cai, Shi-Jie Li, Yun Liu, Yangyan Guo, Tao Li, and
  Ming-Ming Cheng.
\newblock Cubemapslam: A piecewise-pinhole monocular fisheye slam system.
\newblock In {\em Asian Conference on Computer Vision}, pages 34--49. Springer,
  2018.

\bibitem{mur2017visual}
Ra{\'u}l Mur-Artal and Juan~D Tard{\'o}s.
\newblock Visual-inertial monocular slam with map reuse.
\newblock {\em IEEE Robotics and Automation Letters}, 2(2):796--803, 2017.

\bibitem{forster2016manifold}
Christian Forster, Luca Carlone, Frank Dellaert, and Davide Scaramuzza.
\newblock On-manifold preintegration for real-time visual--inertial odometry.
\newblock {\em IEEE Transactions on Robotics}, 33(1):1--21, 2016.

\bibitem{2018-schlegel-proslam}
D.~Schlegel, M.~Colosi, and G.~Grisetti.
\newblock {ProSLAM: Graph SLAM from a Programmer's Perspective}.
\newblock In {\em 2018 IEEE International Conference on Robotics and Automation
  (ICRA)}, pages 1--9, 2018.

\bibitem{zhang2016efficient}
Guofeng Zhang, Haomin Liu, Zilong Dong, Jiaya Jia, Tien-Tsin Wong, and Hujun
  Bao.
\newblock Efficient non-consecutive feature tracking for robust
  structure-from-motion.
\newblock {\em IEEE Transactions on Image Processing}, 25(12):5957--5970, 2016.

\bibitem{openvslam2019}
Shinya Sumikura, Mikiya Shibuya, and Ken Sakurada.
\newblock Openvslam: a versatile visual slam framework, 2019.

\bibitem{pfrommer2019tagslam}
Bernd Pfrommer and Kostas Daniilidis.
\newblock Tagslam: Robust slam with fiducial markers.
\newblock {\em arXiv preprint arXiv:1910.00679}, 2019.

\bibitem{munoz2019ucoslam}
Rafael Munoz-Salinas and Rafael Medina-Carnicer.
\newblock Ucoslam: Simultaneous localization and mapping by fusion of keypoints
  and squared planar markers.
\newblock {\em arXiv preprint arXiv:1902.03729}, 2019.

\bibitem{engel2014lsd}
Jakob Engel, Thomas Sch{\"o}ps, and Daniel Cremers.
\newblock Lsd-slam: Large-scale direct monocular slam.
\newblock In {\em European conference on computer vision}, pages 834--849.
  Springer, 2014.

\bibitem{engel2015large}
Jakob Engel, J{\"o}rg St{\"u}ckler, and Daniel Cremers.
\newblock Large-scale direct slam with stereo cameras.
\newblock In {\em 2015 IEEE/RSJ International Conference on Intelligent Robots
  and Systems (IROS)}, pages 1935--1942. IEEE, 2015.

\bibitem{caruso2015large}
David Caruso, Jakob Engel, and Daniel Cremers.
\newblock Large-scale direct slam for omnidirectional cameras.
\newblock In {\em 2015 IEEE/RSJ International Conference on Intelligent Robots
  and Systems (IROS)}, pages 141--148. IEEE, 2015.

\bibitem{li2018spherical}
Jianfeng Li, Xiaowei Wang, and Shigang Li.
\newblock Spherical-model-based slam on full-view images for indoor
  environments.
\newblock {\em Applied Sciences}, 8(11):2268, 2018.

\bibitem{forster2016svo}
Christian Forster, Zichao Zhang, Michael Gassner, Manuel Werlberger, and Davide
  Scaramuzza.
\newblock Svo: Semidirect visual odometry for monocular and multicamera
  systems.
\newblock {\em IEEE Transactions on Robotics}, 33(2):249--265, 2016.

\bibitem{loo2018cnn}
Shing~Yan Loo, Ali~Jahani Amiri, Syamsiah Mashohor, Sai~Hong Tang, and Hong
  Zhang.
\newblock Cnn-svo: Improving the mapping in semi-direct visual odometry using
  single-image depth prediction.
\newblock {\em arXiv preprint arXiv:1810.01011}, 2018.

\bibitem{DBLP:journals/corr/EngelKC16}
Jakob Engel, Vladlen Koltun, and Daniel Cremers.
\newblock Direct sparse odometry.
\newblock {\em CoRR}, abs/1607.02565, 2016.

\bibitem{engel2017direct}
Jakob Engel, Vladlen Koltun, and Daniel Cremers.
\newblock Direct sparse odometry.
\newblock {\em IEEE transactions on pattern analysis and machine intelligence},
  40(3):611--625, 2017.

\bibitem{rebecq2016evo}
Henri Rebecq, Timo Horstsch{\"a}fer, Guillermo Gallego, and Davide Scaramuzza.
\newblock Evo: A geometric approach to event-based 6-dof parallel tracking and
  mapping in real time.
\newblock {\em IEEE Robotics and Automation Letters}, 2(2):593--600, 2016.

\bibitem{zhou2018semi}
Yi~Zhou, Guillermo Gallego, Henri Rebecq, Laurent Kneip, Hongdong Li, and
  Davide Scaramuzza.
\newblock Semi-dense 3d reconstruction with a stereo event camera.
\newblock In {\em Proceedings of the European Conference on Computer Vision
  (ECCV)}, pages 235--251, 2018.

\bibitem{weikersdorfer2013simultaneous}
David Weikersdorfer, Raoul Hoffmann, and J{\"o}rg Conradt.
\newblock Simultaneous localization and mapping for event-based vision systems.
\newblock In {\em International Conference on Computer Vision Systems}, pages
  133--142. Springer, 2013.

\bibitem{weikersdorfer2014event}
David Weikersdorfer, David~B Adrian, Daniel Cremers, and J{\"o}rg Conradt.
\newblock Event-based 3d slam with a depth-augmented dynamic vision sensor.
\newblock In {\em 2014 IEEE International Conference on Robotics and Automation
  (ICRA)}, pages 359--364. IEEE, 2014.

\bibitem{civera2008inverse}
Javier Civera, Andrew~J Davison, and JM~Martinez Montiel.
\newblock Inverse depth parametrization for monocular slam.
\newblock {\em IEEE transactions on robotics}, 24(5):932--945, 2008.

\bibitem{newcombe2011dtam}
Richard~A Newcombe, Steven~J Lovegrove, and Andrew~J Davison.
\newblock Dtam: Dense tracking and mapping in real-time.
\newblock In {\em 2011 international conference on computer vision}, pages
  2320--2327. IEEE, 2011.

\bibitem{greene2016multi}
W~Nicholas Greene, Kyel Ok, Peter Lommel, and Nicholas Roy.
\newblock Multi-level mapping: Real-time dense monocular slam.
\newblock In {\em 2016 IEEE International Conference on Robotics and Automation
  (ICRA)}, pages 833--840. IEEE, 2016.

\bibitem{newcombe2011kinectfusion}
Richard~A Newcombe, Shahram Izadi, Otmar Hilliges, David Molyneaux, David Kim,
  Andrew~J Davison, Pushmeet Kohli, Jamie Shotton, Steve Hodges, and Andrew~W
  Fitzgibbon.
\newblock Kinectfusion: Real-time dense surface mapping and tracking.
\newblock In {\em ISMAR}, volume~11, pages 127--136, 2011.

\bibitem{izadi2011kinectfusion}
Shahram Izadi, David Kim, Otmar Hilliges, David Molyneaux, Richard Newcombe,
  Pushmeet Kohli, Jamie Shotton, Steve Hodges, Dustin Freeman, Andrew Davison,
  et~al.
\newblock Kinectfusion: real-time 3d reconstruction and interaction using a
  moving depth camera.
\newblock In {\em Proceedings of the 24th annual ACM symposium on User
  interface software and technology}, pages 559--568. ACM, 2011.

\bibitem{steinbrucker2011real}
Frank Steinbr{\"u}cker, J{\"u}rgen Sturm, and Daniel Cremers.
\newblock Real-time visual odometry from dense rgb-d images.
\newblock In {\em 2011 IEEE International Conference on Computer Vision
  Workshops (ICCV Workshops)}, pages 719--722. IEEE, 2011.

\bibitem{kerl2013robust}
Christian Kerl, J{\"u}rgen Sturm, and Daniel Cremers.
\newblock Robust odometry estimation for rgb-d cameras.
\newblock In {\em 2013 IEEE International Conference on Robotics and
  Automation}, pages 3748--3754. IEEE, 2013.

\bibitem{kerl2013dense}
Christian Kerl, J{\"u}rgen Sturm, and Daniel Cremers.
\newblock Dense visual slam for rgb-d cameras.
\newblock In {\em 2013 IEEE/RSJ International Conference on Intelligent Robots
  and Systems}, pages 2100--2106. IEEE, 2013.

\bibitem{endres20133}
Felix Endres, J{\"u}rgen Hess, J{\"u}rgen Sturm, Daniel Cremers, and Wolfram
  Burgard.
\newblock 3-d mapping with an rgb-d camera.
\newblock {\em IEEE transactions on robotics}, 30(1):177--187, 2013.

\bibitem{whelan2012kintinuous}
Thomas Whelan, Michael Kaess, Maurice Fallon, Hordur Johannsson, John~J
  Leonard, and John McDonald.
\newblock Kintinuous: Spatially extended kinectfusion.
\newblock 2012.

\bibitem{whelan2015real}
Thomas Whelan, Michael Kaess, Hordur Johannsson, Maurice Fallon, John~J
  Leonard, and John McDonald.
\newblock Real-time large-scale dense rgb-d slam with volumetric fusion.
\newblock {\em The International Journal of Robotics Research},
  34(4-5):598--626, 2015.

\bibitem{Whelan2011Robust}
Thomas Whelan, Hordur Johannsson, Michael Kaess, John~J. Leonard, and John
  Mcdonald.
\newblock Robust real-time visual odometry for dense rgb-d mapping.
\newblock In {\em IEEE International Conference on Robotics and Automation},
  2011.

\bibitem{labbe2014online}
Mathieu Labbe and Fran{\c{c}}ois Michaud.
\newblock Online global loop closure detection for large-scale multi-session
  graph-based slam.
\newblock In {\em 2014 IEEE/RSJ International Conference on Intelligent Robots
  and Systems}, pages 2661--2666. IEEE, 2014.

\bibitem{labbeappearance}
MM~Labb{\'e} and F~Michaud.
\newblock Appearance-based loop closure detection in real-time for large-scale
  and long-term operation.
\newblock {\em IEEE Transactions on Robotics}, pages 734--745.

\bibitem{labbe2011memory}
Mathieu Labb{\'e} and Fran{\c{c}}ois Michaud.
\newblock Memory management for real-time appearance-based loop closure
  detection.
\newblock In {\em 2011 IEEE/RSJ International Conference on Intelligent Robots
  and Systems}, pages 1271--1276. IEEE, 2011.

\bibitem{labbe2019rtab}
Mathieu Labb{\'e} and Fran{\c{c}}ois Michaud.
\newblock Rtab-map as an open-source lidar and visual simultaneous localization
  and mapping library for large-scale and long-term online operation.
\newblock {\em Journal of Field Robotics}, 36(2):416--446, 2019.

\bibitem{newcombe2015dynamicfusion}
Richard~A Newcombe, Dieter Fox, and Steven~M Seitz.
\newblock Dynamicfusion: Reconstruction and tracking of non-rigid scenes in
  real-time.
\newblock In {\em Proceedings of the IEEE conference on computer vision and
  pattern recognition}, pages 343--352, 2015.

\bibitem{innmann2016volumedeform}
Matthias Innmann, Michael Zollh{\"o}fer, Matthias Nie{\ss}ner, Christian
  Theobalt, and Marc Stamminger.
\newblock Volumedeform: Real-time volumetric non-rigid reconstruction.
\newblock In {\em European Conference on Computer Vision}, pages 362--379.
  Springer, 2016.

\bibitem{dou2016fusion4d}
Mingsong Dou, Sameh Khamis, Yury Degtyarev, Philip Davidson, Sean~Ryan Fanello,
  Adarsh Kowdle, Sergio~Orts Escolano, Christoph Rhemann, David Kim, Jonathan
  Taylor, et~al.
\newblock Fusion4d: Real-time performance capture of challenging scenes.
\newblock {\em ACM Transactions on Graphics (TOG)}, 35(4):114, 2016.

\bibitem{whelan2015elasticfusion}
Thomas Whelan, Stefan Leutenegger, R~Salas-Moreno, Ben Glocker, and Andrew
  Davison.
\newblock Elasticfusion: Dense slam without a pose graph.
\newblock Robotics: Science and Systems, 2015.

\bibitem{whelan2016elasticfusion}
Thomas Whelan, Renato~F Salas-Moreno, Ben Glocker, Andrew~J Davison, and Stefan
  Leutenegger.
\newblock Elasticfusion: Real-time dense slam and light source estimation.
\newblock {\em The International Journal of Robotics Research},
  35(14):1697--1716, 2016.

\bibitem{InfiniTAM_arXiv_2017}
V~A Prisacariu, O~K{\"a}hler, S~Golodetz, M~Sapienza, T~Cavallari, P~H~S Torr,
  and D~W Murray.
\newblock {InfiniTAM v3: A Framework for Large-Scale 3D Reconstruction with
  Loop Closure}.
\newblock {\em arXiv pre-print arXiv:1708.00783v1}, 2017.

\bibitem{InfiniTAM_ECCV_2016}
Olaf K{\"{a}}hler, Victor~Adrian Prisacariu, and David~W. Murray.
\newblock Real-time large-scale dense 3d reconstruction with loop closure.
\newblock In {\em Computer Vision - {ECCV} 2016 - 14th European Conference,
  Amsterdam, The Netherlands, October 11-14, 2016, Proceedings, Part {VIII}},
  pages 500--516, 2016.

\bibitem{dai2017bundlefusion}
Angela Dai, Matthias Nie{\ss}ner, Michael Zoll{\"o}fer, Shahram Izadi, and
  Christian Theobalt.
\newblock Bundlefusion: Real-time globally consistent 3d reconstruction using
  on-the-fly surface re-integration.
\newblock {\em ACM Transactions on Graphics 2017 (TOG)}, 2017.

\bibitem{houseago2019ko}
Charlie Houseago, Michael Bloesch, and Stefan Leutenegger.
\newblock Ko-fusion: Dense visual slam with tightly-coupled kinematic and
  odometric tracking.
\newblock In {\em 2019 International Conference on Robotics and Automation
  (ICRA)}, pages 4054--4060. IEEE, 2019.

\bibitem{cvivsic2017soft}
Igor Cvi{\v{s}}ic, Josip Cesic, Ivan Markovic, and Ivan Petrovic.
\newblock Soft-slam: Computationally efficient stereo visual slam for
  autonomous uavs.
\newblock {\em Journal of field robotics}, 2017.

\bibitem{cvivsic2015stereo}
Igor Cvi{\v{s}}i{\'c} and Ivan Petrovi{\'c}.
\newblock Stereo odometry based on careful feature selection and tracking.
\newblock In {\em 2015 European Conference on Mobile Robots (ECMR)}, pages
  1--6. IEEE, 2015.

\bibitem{liu2017robust}
Haomin Liu, Chen Li, Guojun Chen, Guofeng Zhang, Michael Kaess, and Hujun Bao.
\newblock Robust keyframe-based dense slam with an rgb-d camera.
\newblock {\em arXiv preprint arXiv:1711.05166}, 2017.

\bibitem{dai2018rgb}
Weichen Dai, Yu~Zhang, Ping Li, and Zheng Fang.
\newblock Rgb-d slam in dynamic environments using points correlations.
\newblock {\em arXiv preprint arXiv:1811.03217}, 2018.

\bibitem{schneider2018maplab}
T.~Schneider, M.~T. Dymczyk, M.~Fehr, K.~Egger, S.~Lynen, I.~Gilitschenski, and
  R.~Siegwart.
\newblock maplab: An open framework for research in visual-inertial mapping and
  localization.
\newblock {\em IEEE Robotics and Automation Letters}, 2018.

\bibitem{2019PointMVSNet}
Jing Xu Hao~Su Rui~Chen, Songfang~Han.
\newblock Point-based multi-view stereo network.
\newblock {\em arXiv preprint arXiv:1908.04422}, 2019.

\bibitem{xu2018mid}
Binbin Xu, Wenbin Li, Dimos Tzoumanikas, Michael Bloesch, Andrew Davison, and
  Stefan Leutenegger.
\newblock Mid-fusion: Octree-based object-level multi-instance dynamic slam.
\newblock {\em arXiv preprint arXiv:1812.07976}, 2018.

\bibitem{8613746}
M.~Runz, M.~Buffier, and L.~Agapito.
\newblock Maskfusion: Real-time recognition, tracking and reconstruction of
  multiple moving objects.
\newblock In {\em 2018 IEEE International Symposium on Mixed and Augmented
  Reality (ISMAR)}, pages 10--20, Oct 2018.

\bibitem{strasdat2010scale}
Hauke Strasdat, J~Montiel, and Andrew~J Davison.
\newblock Scale drift-aware large scale monocular slam.
\newblock {\em Robotics: Science and Systems VI}, 2(3):7, 2010.

\bibitem{leutenegger2015keyframe}
Stefan Leutenegger, Simon Lynen, Michael Bosse, Roland Siegwart, and Paul
  Furgale.
\newblock Keyframe-based visual--inertial odometry using nonlinear
  optimization.
\newblock {\em The International Journal of Robotics Research}, 34(3):314--334,
  2015.

\bibitem{huang2014towards}
Guoquan Huang, Michael Kaess, and John~J Leonard.
\newblock Towards consistent visual-inertial navigation.
\newblock In {\em 2014 IEEE International Conference on Robotics and Automation
  (ICRA)}, pages 4926--4933. IEEE, 2014.

\bibitem{li2013high}
Mingyang Li and Anastasios~I Mourikis.
\newblock High-precision, consistent ekf-based visual-inertial odometry.
\newblock {\em The International Journal of Robotics Research}, 32(6):690--711,
  2013.

\bibitem{Campos2019FastAR}
Carlos Campos, J.~M.~M. Montiel, and Juan~D. Tard{\'o}s.
\newblock Fast and robust initialization for visual-inertial slam.
\newblock {\em 2019 International Conference on Robotics and Automation
  (ICRA)}, pages 1288--1294, 2019.

\bibitem{froehlich2017investigation}
Mark Froehlich, Salman Azhar, and Matthew Vanture.
\newblock An investigation of google tango{\textregistered} tablet for low cost
  3d scanning.
\newblock In {\em ISARC. Proceedings of the International Symposium on
  Automation and Robotics in Construction}, volume~34. Vilnius Gediminas
  Technical University, Department of Construction Economics~…, 2017.

\bibitem{garon2016real}
Mathieu Garon, Pierre-Olivier Boulet, Jean-Philippe Doironz, Luc Beaulieu, and
  Jean-Fran{\c{c}}ois Lalonde.
\newblock Real-time high resolution 3d data on the hololens.
\newblock In {\em 2016 IEEE International Symposium on Mixed and Augmented
  Reality (ISMAR-Adjunct)}, pages 189--191. IEEE, 2016.

\bibitem{pfrommer2017penncosyvio}
Bernd Pfrommer, Nitin Sanket, Kostas Daniilidis, and Jonas Cleveland.
\newblock Penncosyvio: A challenging visual inertial odometry benchmark.
\newblock In {\em 2017 IEEE International Conference on Robotics and Automation
  (ICRA)}, pages 3847--3854. IEEE, 2017.

\bibitem{delmerico2018benchmark}
Jeffrey Delmerico and Davide Scaramuzza.
\newblock A benchmark comparison of monocular visual-inertial odometry
  algorithms for flying robots.
\newblock In {\em 2018 IEEE International Conference on Robotics and Automation
  (ICRA)}, pages 2502--2509. IEEE, 2018.

\bibitem{weiss2012vision}
Stephan~M Weiss.
\newblock {\em Vision based navigation for micro helicopters}.
\newblock PhD thesis, ETH Zurich, 2012.

\bibitem{mourikis2007multi}
Anastasios~I Mourikis and Stergios~I Roumeliotis.
\newblock A multi-state constraint kalman filter for vision-aided inertial
  navigation.
\newblock In {\em Proceedings 2007 IEEE International Conference on Robotics
  and Automation}, pages 3565--3572. IEEE, 2007.

\bibitem{sun2018robust}
Ke~Sun, Kartik Mohta, Bernd Pfrommer, Michael Watterson, Sikang Liu, Yash
  Mulgaonkar, Camillo~J Taylor, and Vijay Kumar.
\newblock Robust stereo visual inertial odometry for fast autonomous flight.
\newblock {\em IEEE Robotics and Automation Letters}, 3(2):965--972, 2018.

\bibitem{bloesch2015robust}
Michael Bloesch, Sammy Omari, Marco Hutter, and Roland Siegwart.
\newblock Robust visual inertial odometry using a direct ekf-based approach.
\newblock In {\em 2015 IEEE/RSJ international conference on intelligent robots
  and systems (IROS)}, pages 298--304. IEEE, 2015.

\bibitem{li2017monocular}
Peiliang Li, Tong Qin, Botao Hu, Fengyuan Zhu, and Shaojie Shen.
\newblock Monocular visual-inertial state estimation for mobile augmented
  reality.
\newblock In {\em 2017 IEEE International Symposium on Mixed and Augmented
  Reality (ISMAR)}, pages 11--21. IEEE, 2017.

\bibitem{qin2018online}
Tong Qin and Shaojie Shen.
\newblock Online temporal calibration for monocular visual-inertial systems.
\newblock In {\em 2018 IEEE/RSJ International Conference on Intelligent Robots
  and Systems (IROS)}, pages 3662--3669. IEEE, 2018.

\bibitem{qin2017robust}
Tong Qin and Shaojie Shen.
\newblock Robust initialization of monocular visual-inertial estimation on
  aerial robots.
\newblock In {\em 2017 IEEE/RSJ International Conference on Intelligent Robots
  and Systems (IROS)}, pages 4225--4232. IEEE, 2017.

\bibitem{yang2016monocular}
Zhenfei Yang and Shaojie Shen.
\newblock Monocular visual--inertial state estimation with online
  initialization and camera--imu extrinsic calibration.
\newblock {\em IEEE Transactions on Automation Science and Engineering},
  14(1):39--51, 2016.

\bibitem{liu2018ice}
Haomin Liu, Mingyu Chen, Guofeng Zhang, Hujun Bao, and Yingze Bao.
\newblock Ice-ba: Incremental, consistent and efficient bundle adjustment for
  visual-inertial slam.
\newblock In {\em Proceedings of the IEEE Conference on Computer Vision and
  Pattern Recognition}, pages 1974--1982, 2018.

\bibitem{zou2019structvio}
Danping Zou, Yuanxin Wu, Ling Pei, Haibin Ling, and Wenxian Yu.
\newblock Structvio: Visual-inertial odometry with structural regularity of
  man-made environments.
\newblock {\em IEEE Transactions on Robotics}, 2019.

\bibitem{liu2016robust}
Haomin Liu, Guofeng Zhang, and Hujun Bao.
\newblock Robust keyframe-based monocular slam for augmented reality.
\newblock In {\em 2016 IEEE International Symposium on Mixed and Augmented
  Reality (ISMAR)}, pages 1--10. IEEE, 2016.

\bibitem{mueggler2018continuous}
Elias Mueggler, Guillermo Gallego, Henri Rebecq, and Davide Scaramuzza.
\newblock Continuous-time visual-inertial odometry for event cameras.
\newblock {\em IEEE Transactions on Robotics}, 34(6):1425--1440, 2018.

\bibitem{zhu2017event}
Alex~Zihao Zhu, Nikolay Atanasov, and Kostas Daniilidis.
\newblock Event-based visual inertial odometry.
\newblock In {\em 2017 IEEE Conference on Computer Vision and Pattern
  Recognition (CVPR)}, pages 5816--5824. IEEE, 2017.

\bibitem{nelson2019event}
Kaleb~J Nelson.
\newblock Event-based visual-inertial odometry on a fixed-wing unmanned aerial
  vehicle.
\newblock Technical report, AIR FORCE INSTITUTE OF TECHNOLOGY WRIGHT-PATTERSON
  AFB OH WRIGHT-PATTERSON, 2019.

\bibitem{Eckenhoff2019SensorFailureResilientMV}
Kevin Eckenhoff, Patrick Geneva, and Guoquan Huang.
\newblock Sensor-failure-resilient multi-imu visual-inertial navigation.
\newblock {\em 2019 International Conference on Robotics and Automation
  (ICRA)}, pages 3542--3548, 2019.

\bibitem{shamwell2019unsupervised}
E~Jared Shamwell, Kyle Lindgren, Sarah Leung, and William~D Nothwang.
\newblock Unsupervised deep visual-inertial odometry with online error
  correction for rgb-d imagery.
\newblock {\em IEEE transactions on pattern analysis and machine intelligence},
  2019.

\bibitem{lee2019visual}
Hongyun Lee, Matthew McCrink, and James~W Gregory.
\newblock Visual-inertial odometry for unmanned aerial vehicle using deep
  learning.
\newblock In {\em AIAA Scitech 2019 Forum}, page 1410, 2019.

\bibitem{yang2016pop}
Shichao Yang, Yu~Song, Michael Kaess, and Sebastian Scherer.
\newblock Pop-up slam: Semantic monocular plane slam for low-texture
  environments.
\newblock In {\em 2016 IEEE/RSJ International Conference on Intelligent Robots
  and Systems (IROS)}, pages 1222--1229. IEEE, 2016.

\bibitem{pavlakos20176}
Georgios Pavlakos, Xiaowei Zhou, Aaron Chan, Konstantinos~G Derpanis, and
  Kostas Daniilidis.
\newblock 6-dof object pose from semantic keypoints.
\newblock In {\em 2017 IEEE International Conference on Robotics and Automation
  (ICRA)}, pages 2011--2018. IEEE, 2017.

\bibitem{yi2016lift}
Kwang~Moo Yi, Eduard Trulls, Vincent Lepetit, and Pascal Fua.
\newblock Lift: Learned invariant feature transform.
\newblock In {\em European Conference on Computer Vision}, pages 467--483.
  Springer, 2016.

\bibitem{detone2017toward}
Daniel DeTone, Tomasz Malisiewicz, and Andrew Rabinovich.
\newblock Toward geometric deep slam.
\newblock {\em arXiv preprint arXiv:1707.07410}, 2017.

\bibitem{detone2018superpoint}
Daniel DeTone, Tomasz Malisiewicz, and Andrew Rabinovich.
\newblock Superpoint: Self-supervised interest point detection and description.
\newblock In {\em Proceedings of the IEEE Conference on Computer Vision and
  Pattern Recognition Workshops}, pages 224--236, 2018.

\bibitem{Li2018Stereo}
Peiliang Li, Qin Tong, and Shaojie Shen.
\newblock Stereo vision-based semantic 3d object and ego-motion tracking for
  autonomous driving.
\newblock In {\em European Conference on Computer Vision}, 2018.

\bibitem{tang2019gcnv2}
Jiexiong Tang, Ludvig Ericson, John Folkesson, and Patric Jensfelt.
\newblock Gcnv2: Efficient correspondence prediction for real-time slam.
\newblock {\em arXiv preprint arXiv:1902.11046}, 2019.

\bibitem{grinvald2019volumetric}
Margarita Grinvald, Fadri Furrer, Tonci Novkovic, Jen~Jen Chung, Cesar Cadena,
  Roland Siegwart, and Juan Nieto.
\newblock Volumetric instance-aware semantic mapping and 3d object discovery.
\newblock {\em arXiv preprint arXiv:1903.00268}, 2019.

\bibitem{liang2019salientdso}
Huai-Jen Liang, Nitin~J Sanket, Cornelia Ferm{\"u}ller, and Yiannis Aloimonos.
\newblock Salientdso: Bringing attention to direct sparse odometry.
\newblock {\em IEEE Transactions on Automation Science and Engineering}, 2019.

\bibitem{hosseinzadeh2018structure}
Mehdi Hosseinzadeh, Yasir Latif, Trung Pham, Niko Suenderhauf, and Ian Reid.
\newblock Structure aware slam using quadrics and planes.
\newblock In {\em Asian Conference on Computer Vision}, pages 410--426.
  Springer, 2018.

\bibitem{yang2019cubeslam}
Shichao Yang and Sebastian Scherer.
\newblock Cubeslam: Monocular 3-d object slam.
\newblock {\em IEEE Transactions on Robotics}, 2019.

\bibitem{yang2019monocular}
Shichao Yang and Sebastian Scherer.
\newblock Monocular object and plane slam in structured environments.
\newblock {\em IEEE Robotics and Automation Letters}, 4(4):3145--3152, 2019.

\bibitem{qin2019monogrnet}
Zengyi Qin, Jinglu Wang, and Yan Lu.
\newblock Monogrnet: A geometric reasoning network for 3d object localization.
\newblock {\em The Thirty-Third AAAI Conference on Artificial Intelligence
  (AAAI-19)}, 2019.

\bibitem{Lagorce2013Event}
Xavier Lagorce, Sio~Hoi Ieng, and Ryad Benosman.
\newblock Event-based features for robotic vision.
\newblock In {\em IEEE/RSJ International Conference on Intelligent Robots \&
  Systems}, 2013.

\bibitem{mueggler2017fast}
Elias Mueggler, Chiara Bartolozzi, and Davide Scaramuzza.
\newblock Fast event-based corner detection.
\newblock In {\em BMVC}, 2017.

\bibitem{wu2019recent}
Xiongwei Wu, Doyen Sahoo, and Steven~CH Hoi.
\newblock Recent advances in deep learning for object detection.
\newblock {\em arXiv preprint arXiv:1908.03673}, 2019.

\bibitem{Salas2013SLAM}
Renato~F. Salas-Moreno, Richard~A. Newcombe, Hauke Strasdat, Paul H.~J. Kelly,
  and Andrew~J. Davison.
\newblock Slam++: Simultaneous localisation and mapping at the level of
  objects.
\newblock In {\em Computer Vision \& Pattern Recognition}, 2013.

\bibitem{Li2016Semi}
Xuanpeng Li and Rachid Belaroussi.
\newblock Semi-dense 3d semantic mapping from monocular slam.
\newblock 2016.

\bibitem{mccormac2017semanticfusion}
John McCormac, Ankur Handa, Andrew Davison, and Stefan Leutenegger.
\newblock Semanticfusion: Dense 3d semantic mapping with convolutional neural
  networks.
\newblock In {\em 2017 IEEE International Conference on Robotics and automation
  (ICRA)}, pages 4628--4635. IEEE, 2017.

\bibitem{sunderhauf2017meaningful}
Niko Sunderhauf, Trung~T. Pham, Yasir Latif, Michael Milford, and Ian Reid.
\newblock Meaningful maps with object-oriented semantic mapping.
\newblock In {\em IEEE/RSJ International Conference on Intelligent Robots \&
  Systems}, 2017.

\bibitem{wu2017marrnet}
Jiajun Wu, Yifan Wang, Tianfan Xue, Xingyuan Sun, Bill Freeman, and Josh
  Tenenbaum.
\newblock Marrnet: 3d shape reconstruction via 2.5 d sketches.
\newblock In {\em Advances in neural information processing systems}, pages
  540--550, 2017.

\bibitem{dai20183dmv}
Angela Dai and Matthias Nie{\ss}ner.
\newblock 3dmv: Joint 3d-multi-view prediction for 3d semantic scene
  segmentation.
\newblock In {\em Proceedings of the European Conference on Computer Vision
  (ECCV)}, pages 452--468, 2018.

\bibitem{pix3d}
Xingyuan Sun, Jiajun Wu, Xiuming Zhang, Zhoutong Zhang, Chengkai Zhang, Tianfan
  Xue, Joshua~B Tenenbaum, and William~T Freeman.
\newblock Pix3d: Dataset and methods for single-image 3d shape modeling.
\newblock In {\em IEEE Conference on Computer Vision and Pattern Recognition
  (CVPR)}, 2018.

\bibitem{dai2018scancomplete}
Angela Dai, Daniel Ritchie, Martin Bokeloh, Scott Reed, J{\"u}rgen Sturm, and
  Matthias Nie{\ss}ner.
\newblock Scancomplete: Large-scale scene completion and semantic segmentation
  for 3d scans.
\newblock In {\em Proceedings of the IEEE Conference on Computer Vision and
  Pattern Recognition}, pages 4578--4587, 2018.

\bibitem{McCormac2018FusionVO}
John McCormac, Ronald Clark, Michael Bloesch, Andrew~J. Davison, and Stefan
  Leutenegger.
\newblock Fusion++: Volumetric object-level slam.
\newblock {\em 2018 International Conference on 3D Vision (3DV)}, pages 32--41,
  2018.

\bibitem{dube2018segmap}
Renaud Dub{\'e}, Andrei Cramariuc, Daniel Dugas, Juan Nieto, Roland Siegwart,
  and Cesar Cadena.
\newblock Segmap: 3d segment mapping using data-driven descriptors.
\newblock {\em arXiv preprint arXiv:1804.09557}, 2018.

\bibitem{hou20193d}
Ji~Hou, Angela Dai, and Matthias Nie{\ss}ner.
\newblock 3d-sis: 3d semantic instance segmentation of rgb-d scans.
\newblock In {\em Proceedings of the IEEE Conference on Computer Vision and
  Pattern Recognition}, pages 4421--4430, 2019.

\bibitem{xiang2017rnn}
Yu~Xiang and Dieter Fox.
\newblock Da-rnn: Semantic mapping with data associated recurrent neural
  networks.
\newblock {\em arXiv preprint arXiv:1703.03098}, 2017.

\bibitem{wang2019densefusion}
Chen Wang, Danfei Xu, Yuke Zhu, Roberto Mart{\'\i}n-Mart{\'\i}n, Cewu Lu,
  Li~Fei-Fei, and Silvio Savarese.
\newblock Densefusion: 6d object pose estimation by iterative dense fusion.
\newblock In {\em Proceedings of the IEEE Conference on Computer Vision and
  Pattern Recognition}, pages 3343--3352, 2019.

\bibitem{huang2018ccnet}
Zilong Huang, Xinggang Wang, Lichao Huang, Chang Huang, Yunchao Wei, and Wenyu
  Liu.
\newblock Ccnet: Criss-cross attention for semantic segmentation.
\newblock {\em arXiv preprint arXiv:1811.11721}, 2018.

\bibitem{stromatias2017event}
Evangelos Stromatias, Miguel Soto, Mar{\'\i}a~Teresa Serrano~Gotarredona, and
  Bernab{\'e} Linares~Barranco.
\newblock An event-based classifier for dynamic vision sensor and synthetic
  data.
\newblock {\em Frontiers in Neuroscience, 11 (art{\'\i}culo 360)}, 2017.

\bibitem{maro2018event}
Jean-Matthieu Maro and Ryad Benosman.
\newblock Event-based gesture recognition with dynamic background suppression
  using smartphone computational capabilities.
\newblock {\em arXiv preprint arXiv:1811.07802}, 2018.

\bibitem{afshar2018investigation}
Saeed Afshar, Tara~Julia Hamilton, Jonathan~C Tapson, Andr{\'e} van Schaik, and
  Gregory~Kevin Cohen.
\newblock Investigation of event-based memory surfaces for high-speed
  detection, unsupervised feature extraction, and object recognition.
\newblock {\em Frontiers in neuroscience}, 12:1047, 2018.

\bibitem{linares2019dynamic}
Alejandro Linares-Barranco, Antonio Rios-Navarro, Ricardo Tapiador-Morales, and
  Tobi Delbruck.
\newblock Dynamic vision sensor integration on fpga-based cnn accelerators for
  high-speed visual classification.
\newblock {\em arXiv preprint arXiv:1905.07419}, 2019.

\bibitem{tateno2017cnn}
Keisuke Tateno, Federico Tombari, Iro Laina, and Nassir Navab.
\newblock Cnn-slam: Real-time dense monocular slam with learned depth
  prediction.
\newblock In {\em Proceedings of the IEEE Conference on Computer Vision and
  Pattern Recognition}, pages 6243--6252, 2017.

\bibitem{mohanty2016deepvo}
Vikram Mohanty, Shubh Agrawal, Shaswat Datta, Arna Ghosh, Vishnu~Dutt Sharma,
  and Debashish Chakravarty.
\newblock Deepvo: A deep learning approach for monocular visual odometry.
\newblock {\em arXiv preprint arXiv:1611.06069}, 2016.

\bibitem{li2019gs3d}
Buyu Li, Wanli Ouyang, Lu~Sheng, Xingyu Zeng, and Xiaogang Wang.
\newblock Gs3d: An efficient 3d object detection framework for autonomous
  driving.
\newblock In {\em Proceedings of the IEEE Conference on Computer Vision and
  Pattern Recognition}, pages 1019--1028, 2019.

\bibitem{li2018undeepvo}
Ruihao Li, Sen Wang, Zhiqiang Long, and Dongbing Gu.
\newblock Undeepvo: Monocular visual odometry through unsupervised deep
  learning.
\newblock In {\em 2018 IEEE International Conference on Robotics and Automation
  (ICRA)}, pages 7286--7291. IEEE, 2018.

\bibitem{li2019learning}
Zhengqi Li, Tali Dekel, Forrester Cole, Richard Tucker, Noah Snavely, Ce~Liu,
  and William~T Freeman.
\newblock Learning the depths of moving people by watching frozen people.
\newblock In {\em Proceedings of the IEEE Conference on Computer Vision and
  Pattern Recognition}, pages 4521--4530, 2019.

\bibitem{8353862}
D.~{Frost}, V.~{Prisacariu}, and D.~{Murray}.
\newblock Recovering stable scale in monocular slam using object-supplemented
  bundle adjustment.
\newblock {\em IEEE Transactions on Robotics}, 34(3):736--747, June 2018.

\bibitem{Sucar2017Bayesian}
Edgar Sucar and Jean~Bernard Hayet.
\newblock Bayesian scale estimation for monocular slam based on generic object
  detection for correcting scale drift.
\newblock 2017.

\bibitem{yin2018geonet}
Zhichao Yin and Jianping Shi.
\newblock Geonet: Unsupervised learning of dense depth, optical flow and camera
  pose.
\newblock In {\em CVPR}, 2018.

\bibitem{bloesch2018codeslam}
Michael Bloesch, Jan Czarnowski, Ronald Clark, Stefan Leutenegger, and Andrew~J
  Davison.
\newblock Codeslam—learning a compact, optimisable representation for dense
  visual slam.
\newblock In {\em Proceedings of the IEEE Conference on Computer Vision and
  Pattern Recognition}, pages 2560--2568, 2018.

\bibitem{brickwedde2018mono}
Fabian Brickwedde, Steffen Abraham, and Rudolf Mester.
\newblock Mono-stixels: monocular depth reconstruction of dynamic street
  scenes.
\newblock In {\em 2018 IEEE International Conference on Robotics and Automation
  (ICRA)}, pages 1--7. IEEE, 2018.

\bibitem{Almalioglu2018GANVOUD}
Yasin Almalioglu, Muhamad Risqi~U. Saputra, Pedro Porto~Buarque de~Gusm{\~a}o,
  Andrew Markham, and Agathoniki Trigoni.
\newblock Ganvo: Unsupervised deep monocular visual odometry and depth
  estimation with generative adversarial networks.
\newblock {\em 2019 International Conference on Robotics and Automation
  (ICRA)}, pages 5474--5480, 2018.

\bibitem{chakravarty2019gen}
Punarjay Chakravarty, Praveen Narayanan, and Tom Roussel.
\newblock Gen-slam: Generative modeling for monocular simultaneous localization
  and mapping.
\newblock {\em arXiv preprint arXiv:1902.02086}, 2019.

\bibitem{Lasinger2019}
Katrin Lasinger, Ren\'{e} Ranftl, Konrad Schindler, and Vladlen Koltun.
\newblock Towards robust monocular depth estimation: Mixing datasets for
  zero-shot cross-dataset transfer.
\newblock {\em arXiv:1907.01341}, 2019.

\bibitem{huang2018deepmvs}
Po-Han Huang, Kevin Matzen, Johannes Kopf, Narendra Ahuja, and Jia-Bin Huang.
\newblock Deepmvs: Learning multi-view stereopsis.
\newblock In {\em Proceedings of the IEEE Conference on Computer Vision and
  Pattern Recognition}, pages 2821--2830, 2018.

\bibitem{DeepV2D}
Jia~Deng Zachary~Teed.
\newblock Deepv2d: Video to depth with differentiable structure from motion.
\newblock {\em arXiv:1812.04605}, 2018.

\bibitem{haessig2019spiking}
Germain Haessig, Xavier Berthelon, Sio-Hoi Ieng, and Ryad Benosman.
\newblock A spiking neural network model of depth from defocus for event-based
  neuromorphic vision.
\newblock {\em Scientific reports}, 9(1):3744, 2019.

\bibitem{gallego2018unifying}
Guillermo Gallego, Henri Rebecq, and Davide Scaramuzza.
\newblock A unifying contrast maximization framework for event cameras, with
  applications to motion, depth, and optical flow estimation.
\newblock In {\em Proceedings of the IEEE Conference on Computer Vision and
  Pattern Recognition}, pages 3867--3876, 2018.

\bibitem{xie2017event}
Zhen Xie, Shengyong Chen, and Garrick Orchard.
\newblock Event-based stereo depth estimation using belief propagation.
\newblock {\em Frontiers in neuroscience}, 11:535, 2017.

\bibitem{konda2015learning}
Kishore~Reddy Konda and Roland Memisevic.
\newblock Learning visual odometry with a convolutional network.
\newblock In {\em VISAPP (1)}, pages 486--490, 2015.

\bibitem{costante2015exploring}
Gabriele Costante, Michele Mancini, Paolo Valigi, and Thomas~A Ciarfuglia.
\newblock Exploring representation learning with cnns for frame-to-frame
  ego-motion estimation.
\newblock {\em IEEE robotics and automation letters}, 1(1):18--25, 2015.

\bibitem{kendall2015posenet}
Alex Kendall, Matthew Grimes, and Roberto Cipolla.
\newblock Posenet: A convolutional network for real-time 6-dof camera
  relocalization.
\newblock In {\em Proceedings of the IEEE international conference on computer
  vision}, pages 2938--2946, 2015.

\bibitem{clark2017vinet}
Ronald Clark, Sen Wang, Hongkai Wen, Andrew Markham, and Niki Trigoni.
\newblock Vinet: Visual-inertial odometry as a sequence-to-sequence learning
  problem.
\newblock In {\em Thirty-First AAAI Conference on Artificial Intelligence},
  2017.

\bibitem{wang2017deepvo}
Sen Wang, Ronald Clark, Hongkai Wen, and Niki Trigoni.
\newblock Deepvo: Towards end-to-end visual odometry with deep recurrent
  convolutional neural networks.
\newblock In {\em 2017 IEEE International Conference on Robotics and Automation
  (ICRA)}, pages 2043--2050. IEEE, 2017.

\bibitem{zhou2017unsupervised}
Tinghui Zhou, Matthew Brown, Noah Snavely, and David~G Lowe.
\newblock Unsupervised learning of depth and ego-motion from video.
\newblock In {\em Proceedings of the IEEE Conference on Computer Vision and
  Pattern Recognition}, pages 1851--1858, 2017.

\bibitem{vijayanarasimhan2017sfm}
Sudheendra Vijayanarasimhan, Susanna Ricco, Cordelia Schmid, Rahul Sukthankar,
  and Katerina Fragkiadaki.
\newblock Sfm-net: Learning of structure and motion from video.
\newblock {\em arXiv preprint arXiv:1704.07804}, 2017.

\bibitem{Lianos2018VSO}
Konstantinos~Nektarios Lianos, Johannes~L. Schönberger, Marc Pollefeys, and
  Torsten Sattler.
\newblock Vso: Visual semantic odometry.
\newblock In {\em European Conference on Computer Vision (ECCV)}, 2018.

\bibitem{clark2017vidloc}
Ronald Clark, Sen Wang, Andrew Markham, Niki Trigoni, and Hongkai Wen.
\newblock Vidloc: A deep spatio-temporal model for 6-dof video-clip
  relocalization.
\newblock In {\em Proceedings of the IEEE Conference on Computer Vision and
  Pattern Recognition}, pages 6856--6864, 2017.

\bibitem{gallego2015event}
Guillermo Gallego, Christian Forster, Elias Mueggler, and Davide Scaramuzza.
\newblock Event-based camera pose tracking using a generative event model.
\newblock {\em arXiv preprint arXiv:1510.01972}, 2015.

\bibitem{reverter2016neuromorphic}
David Reverter~Valeiras, Garrick Orchard, Sio-Hoi Ieng, and Ryad~B Benosman.
\newblock Neuromorphic event-based 3d pose estimation.
\newblock {\em Frontiers in neuroscience}, 9:522, 2016.

\bibitem{Bowman2017Probabilistic}
Sean~L. Bowman, Nikolay Atanasov, Kostas Daniilidis, and George~J. Pappas.
\newblock Probabilistic data association for semantic slam.
\newblock In {\em IEEE International Conference on Robotics \& Automation},
  2017.

\bibitem{merrill2018lightweight}
Nate Merrill and Guoquan Huang.
\newblock Lightweight unsupervised deep loop closure.
\newblock {\em arXiv preprint arXiv:1805.07703}, 2018.

\bibitem{Stenborg2018Long}
Erik Stenborg, Carl Toft, and Lars Hammarstrand.
\newblock Long-term visual localization using semantically segmented images.
\newblock pages 6484--6490, 2018.

\bibitem{gawel2018x}
Abel Gawel, Carlo Del~Don, Roland Siegwart, Juan Nieto, and Cesar Cadena.
\newblock X-view: Graph-based semantic multi-view localization.
\newblock {\em IEEE Robotics and Automation Letters}, 3(3):1687--1694, 2018.

\bibitem{doherty2019multimodal}
Kevin Doherty, Dehann Fourie, and John Leonard.
\newblock Multimodal semantic slam with probabilistic data association.
\newblock In {\em 2019 IEEE International Conference on Robotics and Automation
  (ICRA). IEEE}, 2019.

\bibitem{censi2013low}
Andrea Censi, Jonas Strubel, Christian Brandli, Tobi Delbruck, and Davide
  Scaramuzza.
\newblock Low-latency localization by active led markers tracking using a
  dynamic vision sensor.
\newblock In {\em 2013 IEEE/RSJ International Conference on Intelligent Robots
  and Systems}, pages 891--898. IEEE, 2013.

\bibitem{tan2013robust}
Wei Tan, Haomin Liu, Zilong Dong, Guofeng Zhang, and Hujun Bao.
\newblock Robust monocular slam in dynamic environments.
\newblock In {\em 2013 IEEE International Symposium on Mixed and Augmented
  Reality (ISMAR)}, pages 209--218. IEEE, 2013.

\bibitem{yu2018ds}
Chao Yu, Zuxin Liu, Xin-Jun Liu, Fugui Xie, Yi~Yang, Qi~Wei, and Qiao Fei.
\newblock Ds-slam: A semantic visual slam towards dynamic environments.
\newblock In {\em 2018 IEEE/RSJ International Conference on Intelligent Robots
  and Systems (IROS)}, pages 1168--1174. IEEE, 2018.

\bibitem{matterport_maskrcnn_2017}
Waleed Abdulla.
\newblock Mask r-cnn for object detection and instance segmentation on keras
  and tensorflow, 2017.

\bibitem{runz2017co}
Martin R{\"u}nz and Lourdes Agapito.
\newblock Co-fusion: Real-time segmentation, tracking and fusion of multiple
  objects.
\newblock In {\em 2017 IEEE International Conference on Robotics and Automation
  (ICRA)}, pages 4471--4478. IEEE, 2017.

\bibitem{Zhong2018Detect}
Fangwei Zhong, Wang Sheng, Ziqi Zhang, China Chen, and Yizhou Wang.
\newblock Detect-slam: Making object detection and slam mutually beneficial.
\newblock In {\em IEEE Winter Conference on Applications of Computer Vision},
  2018.

\bibitem{bescos2018dynaslam}
Berta Bescos, Jos{\'e}~M F{\'a}cil, Javier Civera, and Jos{\'e} Neira.
\newblock Dynaslam: Tracking, mapping, and inpainting in dynamic scenes.
\newblock {\em IEEE Robotics and Automation Letters}, 3(4):4076--4083, 2018.

\bibitem{scona2018staticfusion}
Raluca Scona, Mariano Jaimez, Yvan~R Petillot, Maurice Fallon, and Daniel
  Cremers.
\newblock Staticfusion: Background reconstruction for dense rgb-d slam in
  dynamic environments.
\newblock In {\em 2018 IEEE International Conference on Robotics and Automation
  (ICRA)}, pages 1--9. IEEE, 2018.

\bibitem{wang2019computationally}
Zemin Wang, Qian Zhang, Jiansheng Li, Shuming Zhang, and Jingbin Liu.
\newblock A computationally efficient semantic slam solution for dynamic
  scenes.
\newblock {\em Remote Sensing}, 11(11):1363, 2019.

\bibitem{xiao2019dynamic}
Linhui Xiao, Jinge Wang, Xiaosong Qiu, Zheng Rong, and Xudong Zou.
\newblock Dynamic-slam: Semantic monocular visual localization and mapping
  based on deep learning in dynamic environment.
\newblock {\em Robotics and Autonomous Systems}, 117:1--16, 2019.

\bibitem{barsan2018robust}
Ioan~Andrei B{\^a}rsan, Peidong Liu, Marc Pollefeys, and Andreas Geiger.
\newblock Robust dense mapping for large-scale dynamic environments.
\newblock In {\em 2018 IEEE International Conference on Robotics and Automation
  (ICRA)}, pages 7510--7517. IEEE, 2018.

\bibitem{kim2019simvodis}
Se-Ho~Kim Ue-Hwan~Kim and Jong-Hwan Kim.
\newblock Simvodis: Simultaneous visual odometry, object detection, and
  instance segmentation.
\newblock {\em IEEE Transactions on Pattern Analysis and Machine Intelligence,
  Under Review}, 2019.

\bibitem{sualeh2019simultaneous}
Muhammad Sualeh and Gon-Woo Kim.
\newblock Simultaneous localization and mapping in the epoch of semantics: A
  survey.
\newblock {\em International Journal of Control, Automation and Systems},
  17(3):729--742, 2019.

\bibitem{wang2018dels}
Peng Wang, Ruigang Yang, Binbin Cao, Wei Xu, and Yuanqing Lin.
\newblock Dels-3d: Deep localization and segmentation with a 3d semantic map.
\newblock In {\em Proceedings of the IEEE Conference on Computer Vision and
  Pattern Recognition}, pages 5860--5869, 2018.

\bibitem{atanasov2018unifying}
Nikolay Atanasov, Sean~L Bowman, Kostas Daniilidis, and George~J Pappas.
\newblock A unifying view of geometry, semantics, and data association in slam.
\newblock In {\em IJCAI}, pages 5204--5208, 2018.

\bibitem{gomez2019pl}
Ruben Gomez-Ojeda, Francisco-Angel Moreno, David Zu{\~n}iga-No{\"e}l, Davide
  Scaramuzza, and Javier Gonzalez-Jimenez.
\newblock Pl-slam: a stereo slam system through the combination of points and
  line segments.
\newblock {\em IEEE Transactions on Robotics}, 2019.

\bibitem{zhou2015structslam}
Huizhong Zhou, Danping Zou, Ling Pei, Rendong Ying, Peilin Liu, and Wenxian Yu.
\newblock Structslam: Visual slam with building structure lines.
\newblock {\em IEEE Transactions on Vehicular Technology}, 64(4):1364--1375,
  2015.

\bibitem{riisgaard2005dummies}
Søren Riisgaard and Morten~Rufus Blas.
\newblock Slam for dummies: A tutorial approach to simultaneous localization
  and mapping.
\newblock Technical report, 2005.

\bibitem{rehder2016extending}
Joern Rehder, Janosch Nikolic, Thomas Schneider, Timo Hinzmann, and Roland
  Siegwart.
\newblock Extending kalibr: Calibrating the extrinsics of multiple imus and of
  individual axes.
\newblock In {\em 2016 IEEE International Conference on Robotics and Automation
  (ICRA)}, pages 4304--4311. IEEE, 2016.

\bibitem{Eckenhoff2019MultiCameraVN}
Kevin Eckenhoff, Patrick Geneva, Jesse Bloecker, and Guoquan Huang.
\newblock Multi-camera visual-inertial navigation with online intrinsic and
  extrinsic calibration.
\newblock {\em 2019 International Conference on Robotics and Automation
  (ICRA)}, pages 3158--3164, 2019.

\bibitem{tpm_icra2014}
A.~Tedaldi, A.~Pretto, and E.~Menegatti.
\newblock A robust and easy to implement method for imu calibration without
  external equipments.
\newblock In {\em Proc. of: IEEE International Conference on Robotics and
  Automation (ICRA)}, pages 3042--3049, 2014.

\bibitem{pg_imeko2014}
A.~Pretto and G.~Grisetti.
\newblock Calibration and performance evaluation of low-cost imus.
\newblock In {\em Proc. of: 20th IMEKO TC4 International Symposium}, pages
  429--434, 2014.

\bibitem{li2014high}
Mingyang Li, Hongsheng Yu, Xing Zheng, and Anastasios~I Mourikis.
\newblock High-fidelity sensor modeling and self-calibration in vision-aided
  inertial navigation.
\newblock In {\em 2014 IEEE International Conference on Robotics and Automation
  (ICRA)}, pages 409--416. IEEE, 2014.

\bibitem{chen2019selective}
Changhao Chen, Stefano Rosa, Yishu Miao, Chris~Xiaoxuan Lu, Wei Wu, Andrew
  Markham, and Niki Trigoni.
\newblock Selective sensor fusion for neural visual-inertial odometry.
\newblock In {\em Proceedings of the IEEE Conference on Computer Vision and
  Pattern Recognition}, pages 10542--10551, 2019.

\bibitem{Schops_2019_CVPR}
Thomas Schops, Torsten Sattler, and Marc Pollefeys.
\newblock Bad slam: Bundle adjusted direct rgb-d slam.
\newblock In {\em The IEEE Conference on Computer Vision and Pattern
  Recognition (CVPR)}, June 2019.

\bibitem{harmat2015multi}
Adam Harmat, Michael Trentini, and Inna Sharf.
\newblock Multi-camera tracking and mapping for unmanned aerial vehicles in
  unstructured environments.
\newblock {\em Journal of Intelligent \& Robotic Systems}, 78(2):291--317,
  2015.

\bibitem{UrbanMultiColSLAM16}
Steffen Urban and Stefan Hinz.
\newblock {MultiCol-SLAM} - a modular real-time multi-camera slam system.
\newblock {\em arXiv preprint arXiv:1610.07336}, 2016.

\bibitem{bloesch2017iterated}
Michael Bloesch, Michael Burri, Sammy Omari, Marco Hutter, and Roland Siegwart.
\newblock Iterated extended kalman filter based visual-inertial odometry using
  direct photometric feedback.
\newblock {\em The International Journal of Robotics Research},
  36(10):1053--1072, 2017.

\bibitem{ye2019tightly}
Haoyang Ye, Yuying Chen, and Ming Liu.
\newblock Tightly coupled 3d lidar inertial odometry and mapping.
\newblock {\em arXiv preprint arXiv:1904.06993}, 2019.

\bibitem{yin2018extrinsic}
Deyu Yin, Jingbin Liu, Teng Wu, Keke Liu, Juha Hyypp{\"a}, and Ruizhi Chen.
\newblock Extrinsic calibration of 2d laser rangefinders using an existing
  cuboid-shaped corridor as the reference.
\newblock {\em Sensors}, 18(12):4371, 2018.

\bibitem{chen2018extrinsic}
Shoubin Chen, Jingbin Liu, Teng Wu, Wenchao Huang, Keke Liu, Deyu Yin, Xinlian
  Liang, Juha Hyypp{\"a}, and Ruizhi Chen.
\newblock Extrinsic calibration of 2d laser rangefinders based on a mobile
  sphere.
\newblock {\em Remote Sensing}, 10(8):1176, 2018.

\bibitem{levinson2011automatic}
Jesse~Sol Levinson.
\newblock {\em Automatic laser calibration, mapping, and localization for
  autonomous vehicles}.
\newblock Stanford University, 2011.

\bibitem{levinson2013automatic}
Jesse Levinson and Sebastian Thrun.
\newblock Automatic online calibration of cameras and lasers.
\newblock In {\em Robotics: Science and Systems}, volume~2, 2013.

\bibitem{2017arXiv170509785D}
A.~{Dhall}, K.~{Chelani}, V.~{Radhakrishnan}, and K.~M. {Krishna}.
\newblock {LiDAR-Camera Calibration using 3D-3D Point correspondences}.
\newblock {\em ArXiv e-prints}, May 2017.

\bibitem{schneider2017regnet}
Nick Schneider, Florian Piewak, Christoph Stiller, and Uwe Franke.
\newblock Regnet: Multimodal sensor registration using deep neural networks.
\newblock In {\em 2017 IEEE intelligent vehicles symposium (IV)}, pages
  1803--1810. IEEE, 2017.

\bibitem{Graeter2018LIMO}
Johannes Graeter, Alexander Wilczynski, and Martin Lauer.
\newblock Limo: Lidar-monocular visual odometry.
\newblock 2018.

\bibitem{iyer2018calibnet}
Ganesh Iyer, J~Krishna Murthy, K~Madhava Krishna, et~al.
\newblock Calibnet: self-supervised extrinsic calibration using 3d spatial
  transformer networks.
\newblock {\em arXiv preprint arXiv:1803.08181}, 2018.

\bibitem{mirzaei20123d}
Faraz~M Mirzaei, Dimitrios~G Kottas, and Stergios~I Roumeliotis.
\newblock 3d lidar--camera intrinsic and extrinsic calibration: Identifiability
  and analytical least-squares-based initialization.
\newblock {\em The International Journal of Robotics Research}, 31(4):452--467,
  2012.

\bibitem{ishikawa2018lidar}
Ryoichi Ishikawa, Takeshi Oishi, and Katsushi Ikeuchi.
\newblock Lidar and camera calibration using motions estimated by sensor fusion
  odometry.
\newblock In {\em 2018 IEEE/RSJ International Conference on Intelligent Robots
  and Systems (IROS)}, pages 7342--7349. IEEE, 2018.

\bibitem{Levinson2013AutomaticOC}
Jesse Levinson and Sebastian Thrun.
\newblock Automatic online calibration of cameras and lasers.
\newblock In {\em Robotics: Science and Systems}, 2013.

\bibitem{rahmansvin2}
Sharmin Rahman, Alberto~Quattrini Li, and Ioannis Rekleitis.
\newblock Svin2: An underwater slam system using sonar, visual, inertial, and
  depth sensor.

\bibitem{Gu2019EnvironmentDU}
Changjun Gu, Yang Cong, and Gan Sun.
\newblock Environment driven underwater camera-imu calibration for monocular
  visual-inertial slam.
\newblock {\em 2019 International Conference on Robotics and Automation
  (ICRA)}, pages 2405--2411, 2019.

\bibitem{Arain2019ImprovingUO}
Bilal Arain, Chris McCool, Paul Rigby, Daniel Cagara, and Matthew Dunbabin.
\newblock Improving underwater obstacle detection using semantic image
  segmentation.
\newblock {\em 2019 International Conference on Robotics and Automation
  (ICRA)}, pages 9271--9277, 2019.

\bibitem{ferris2007wifi}
Brian Ferris, Dieter Fox, and Neil~D Lawrence.
\newblock Wifi-slam using gaussian process latent variable models.
\newblock In {\em IJCAI}, volume~7, pages 2480--2485, 2007.

\bibitem{aladsani2019leveraging}
Mohammed Aladsani, Ahmed Alkhateeb, and Georgios~C Trichopoulos.
\newblock Leveraging mmwave imaging and communications for simultaneous
  localization and mapping.
\newblock In {\em ICASSP 2019-2019 IEEE International Conference on Acoustics,
  Speech and Signal Processing (ICASSP)}, pages 4539--4543. IEEE, 2019.

\bibitem{kanhere2018position}
Ojas Kanhere and Theodore~S Rappaport.
\newblock Position locationing for millimeter wave systems.
\newblock In {\em 2018 IEEE Global Communications Conference (GLOBECOM)}, pages
  206--212. IEEE, 2018.

\bibitem{rappaport2019wireless}
Theodore~S Rappaport, Yunchou Xing, Ojas Kanhere, Shihao Ju, Arjuna Madanayake,
  Soumyajit Mandal, Ahmed Alkhateeb, and Georgios~C Trichopoulos.
\newblock Wireless communications and applications above 100 ghz: Opportunities
  and challenges for 6g and beyond.
\newblock {\em IEEE Access}, 7:78729--78757, 2019.

\bibitem{Khattak2019KeyframebasedDT}
Shehryar Khattak, Christos Papachristos, and Kostas Alexis.
\newblock Keyframe-based direct thermal–inertial odometry.
\newblock {\em 2019 International Conference on Robotics and Automation
  (ICRA)}, pages 3563--3569, 2019.

\bibitem{ferstl2013image}
David Ferstl, Christian Reinbacher, Rene Ranftl, Matthias R{\"u}ther, and Horst
  Bischof.
\newblock Image guided depth upsampling using anisotropic total generalized
  variation.
\newblock In {\em Proceedings of the IEEE International Conference on Computer
  Vision}, pages 993--1000, 2013.

\bibitem{ku2018defense}
Jason Ku, Ali Harakeh, and Steven~L Waslander.
\newblock In defense of classical image processing: Fast depth completion on
  the cpu.
\newblock In {\em 2018 15th Conference on Computer and Robot Vision (CRV)},
  pages 16--22. IEEE, 2018.

\bibitem{mal2018sparse}
Fangchang Mal and Sertac Karaman.
\newblock Sparse-to-dense: Depth prediction from sparse depth samples and a
  single image.
\newblock In {\em 2018 IEEE International Conference on Robotics and Automation
  (ICRA)}, pages 1--8. IEEE, 2018.

\bibitem{uhrig2017sparsity}
Jonas Uhrig, Nick Schneider, Lukas Schneider, Uwe Franke, Thomas Brox, and
  Andreas Geiger.
\newblock Sparsity invariant cnns.
\newblock In {\em 2017 International Conference on 3D Vision (3DV)}, pages
  11--20. IEEE, 2017.

\bibitem{shivakumar2019dfusenet}
Shreyas~S Shivakumar, Ty~Nguyen, Steven~W Chen, and Camillo~J Taylor.
\newblock Dfusenet: Deep fusion of rgb and sparse depth information for image
  guided dense depth completion.
\newblock {\em arXiv preprint arXiv:1902.00761}, 2019.

\bibitem{chen2018estimating}
Zhao Chen, Vijay Badrinarayanan, Gilad Drozdov, and Andrew Rabinovich.
\newblock Estimating depth from rgb and sparse sensing.
\newblock In {\em Proceedings of the European Conference on Computer Vision
  (ECCV)}, pages 167--182, 2018.

\bibitem{eldesokey2018propagating}
Abdelrahman Eldesokey, Michael Felsberg, and Fahad~Shahbaz Khan.
\newblock Propagating confidences through cnns for sparse data regression.
\newblock {\em arXiv preprint arXiv:1805.11913}, 2018.

\bibitem{zuo2019lic}
Xingxing Zuo, Patrick Geneva, Woosik Lee, Yong Liu, and Guoquan Huang.
\newblock Lic-fusion: Lidar-inertial-camera odometry.
\newblock {\em arXiv preprint arXiv:1909.04102}, 2019.

\bibitem{aycard2011intersection}
Olivier Aycard, Qadeer Baig, Siviu Bota, Fawzi Nashashibi, Sergiu Nedevschi,
  Cosmin Pantilie, Michel Parent, Paulo Resende, and Trung-Dung Vu.
\newblock Intersection safety using lidar and stereo vision sensors.
\newblock In {\em 2011 IEEE Intelligent Vehicles Symposium (IV)}, pages
  863--869. IEEE, 2011.

\bibitem{chavez2015multiple}
Ricardo~Omar Chavez-Garcia and Olivier Aycard.
\newblock Multiple sensor fusion and classification for moving object detection
  and tracking.
\newblock {\em IEEE Transactions on Intelligent Transportation Systems},
  17(2):525--534, 2015.

\bibitem{cho2014multi}
Hyunggi Cho, Young-Woo Seo, BVK~Vijaya Kumar, and Ragunathan~Raj Rajkumar.
\newblock A multi-sensor fusion system for moving object detection and tracking
  in urban driving environments.
\newblock In {\em 2014 IEEE International Conference on Robotics and Automation
  (ICRA)}, pages 1836--1843. IEEE, 2014.

\bibitem{wang2011integrating}
Tao Wang, Nanning Zheng, Jingmin Xin, and Zheng Ma.
\newblock Integrating millimeter wave radar with a monocular vision sensor for
  on-road obstacle detection applications.
\newblock {\em Sensors}, 11(9):8992--9008, 2011.

\bibitem{wan2018robust}
Guowei Wan, Xiaolong Yang, Renlan Cai, Hao Li, Yao Zhou, Hao Wang, and Shiyu
  Song.
\newblock Robust and precise vehicle localization based on multi-sensor fusion
  in diverse city scenes.
\newblock In {\em 2018 IEEE International Conference on Robotics and Automation
  (ICRA)}, pages 4670--4677. IEEE, 2018.

\bibitem{zhang2014real}
Ji~Zhang, Michael Kaess, and Sanjiv Singh.
\newblock Real-time depth enhanced monocular odometry.
\newblock In {\em 2014 IEEE/RSJ International Conference on Intelligent Robots
  and Systems}, pages 4973--4980. IEEE, 2014.

\bibitem{zhang2015visual}
Ji~Zhang and Sanjiv Singh.
\newblock Visual-lidar odometry and mapping: Low-drift, robust, and fast.
\newblock In {\em 2015 IEEE International Conference on Robotics and Automation
  (ICRA)}, pages 2174--2181. IEEE, 2015.

\bibitem{nava2018visual}
Yoshua Nava.
\newblock {\em Visual-LiDAR SLAM with loop closure}.
\newblock PhD thesis, Master’s thesis, KTH Royal Institute of Technology,
  2018.

\bibitem{xu2018slam}
Yinglei Xu, Yongsheng Ou, and Tiantian Xu.
\newblock Slam of robot based on the fusion of vision and lidar.
\newblock In {\em 2018 IEEE International Conference on Cyborg and Bionic
  Systems (CBS)}, pages 121--126. IEEE, 2018.

\bibitem{shao2019stereo}
Weizhao Shao, Srinivasan Vijayarangan, Cong Li, and George Kantor.
\newblock Stereo visual inertial lidar simultaneous localization and mapping.
\newblock {\em arXiv preprint arXiv:1902.10741}, 2019.

\bibitem{andert2015lidar}
Franz Andert, Nikolaus Ammann, and Bolko Maass.
\newblock Lidar-aided camera feature tracking and visual slam for spacecraft
  low-orbit navigation and planetary landing.
\newblock In {\em Advances in Aerospace Guidance, Navigation and Control},
  pages 605--623. Springer, 2015.

\bibitem{xu2018pointfusion}
Danfei Xu, Dragomir Anguelov, and Ashesh Jain.
\newblock Pointfusion: Deep sensor fusion for 3d bounding box estimation.
\newblock In {\em Proceedings of the IEEE Conference on Computer Vision and
  Pattern Recognition}, pages 244--253, 2018.

\bibitem{shin2018roarnet}
Kiwoo Shin, Youngwook~Paul Kwon, and Masayoshi Tomizuka.
\newblock Roarnet: A robust 3d object detection based on region approximation
  refinement.
\newblock {\em arXiv preprint arXiv:1811.03818}, 2018.

\bibitem{ku2018joint}
Jason Ku, Melissa Mozifian, Jungwook Lee, Ali Harakeh, and Steven Waslander.
\newblock Joint 3d proposal generation and object detection from view
  aggregation.
\newblock {\em IROS}, 2018.

\bibitem{hazirbas2016fusenet}
Caner Hazirbas, Lingni Ma, Csaba Domokos, and Daniel Cremers.
\newblock Fusenet: Incorporating depth into semantic segmentation via
  fusion-based cnn architecture.
\newblock In {\em Asian conference on computer vision}, pages 213--228.
  Springer, 2016.

\bibitem{wang2018fusing}
Zining Wang, Wei Zhan, and Masayoshi Tomizuka.
\newblock Fusing bird’s eye view lidar point cloud and front view camera
  image for 3d object detection.
\newblock In {\em 2018 IEEE Intelligent Vehicles Symposium (IV)}, pages 1--6.
  IEEE, 2018.

\bibitem{liang2018deep}
Ming Liang, Bin Yang, Shenlong Wang, and Raquel Urtasun.
\newblock Deep continuous fusion for multi-sensor 3d object detection.
\newblock In {\em Proceedings of the European Conference on Computer Vision
  (ECCV)}, pages 641--656, 2018.

\bibitem{gu20183}
Shuo Gu, Tao Lu, Yigong Zhang, Jose~M Alvarez, Jian Yang, and Hui Kong.
\newblock 3-d lidar+ monocular camera: An inverse-depth-induced fusion
  framework for urban road detection.
\newblock {\em IEEE Transactions on Intelligent Vehicles}, 3(3):351--360, 2018.

\bibitem{cadena2016past}
Cesar Cadena, Luca Carlone, Henry Carrillo, Yasir Latif, Davide Scaramuzza,
  Jos{\'e} Neira, Ian Reid, and John~J Leonard.
\newblock Past, present, and future of simultaneous localization and mapping:
  Toward the robust-perception age.
\newblock {\em IEEE Transactions on robotics}, 32(6):1309--1332, 2016.

\bibitem{cieslewski2018data}
Titus Cieslewski, Siddharth Choudhary, and Davide Scaramuzza.
\newblock Data-efficient decentralized visual slam.
\newblock In {\em 2018 IEEE International Conference on Robotics and Automation
  (ICRA)}, pages 2466--2473. IEEE, 2018.

\bibitem{dwork2011differential}
Cynthia Dwork.
\newblock Differential privacy.
\newblock {\em Encyclopedia of Cryptography and Security}, pages 338--340,
  2011.

\bibitem{mcsherry2007mechanism}
Frank McSherry and Kunal Talwar.
\newblock Mechanism design via differential privacy.
\newblock In {\em FOCS}, volume~7, pages 94--103, 2007.

\bibitem{sarieddeen2019next}
Hadi Sarieddeen, Nasir Saeed, Tareq~Y Al-Naffouri, and Mohamed-Slim Alouini.
\newblock Next generation terahertz communications: A rendezvous of sensing,
  imaging and localization.
\newblock {\em arXiv preprint arXiv:1909.10462}, 2019.

\bibitem{letaief2019roadmap}
Khaled~B Letaief, Wei Chen, Yuanming Shi, Jun Zhang, and Ying-Jun~Angela Zhang.
\newblock The roadmap to 6g: Ai empowered wireless networks.
\newblock {\em IEEE Communications Magazine}, 57(8):84--90, 2019.

\bibitem{saad2019vision}
Walid Saad, Mehdi Bennis, and Mingzhe Chen.
\newblock A vision of 6g wireless systems: Applications, trends, technologies,
  and open research problems.
\newblock {\em IEEE network}, 2019.

\bibitem{wymeersch2019radio}
Henk Wymeersch, Jiguang He, Beno{\^\i}t Denis, Antonio Clemente, and Markku
  Juntti.
\newblock Radio localization and mapping with reconfigurable intelligent
  surfaces.
\newblock {\em arXiv preprint arXiv:1912.09401}, 2019.

\bibitem{giordani2019towards}
Marco Giordani, Michele Polese, Marco Mezzavilla, Sundeep Rangan, and Michele
  Zorzi.
\newblock Towards 6g networks: Use cases and technologies.
\newblock {\em arXiv preprint arXiv:1903.12216}, 2019.

\end{thebibliography}

\end{document}